\documentclass[letterpaper]{article} % DO NOT CHANGE THIS
\usepackage{aaai25}  % DO NOT CHANGE THIS
\usepackage{times}  % DO NOT CHANGE THIS
\usepackage{helvet}  % DO NOT CHANGE THIS
\usepackage{courier}  % DO NOT CHANGE THIS
\usepackage[hyphens]{url}  % DO NOT CHANGE THIS
\usepackage{graphicx} % DO NOT CHANGE THIS
\usepackage{natbib}  % DO NOT CHANGE THIS AND DO NOT ADD ANY OPTIONS TO IT
\usepackage{caption} % DO NOT CHANGE THIS AND DO NOT ADD ANY OPTIONS TO IT
\usepackage{algorithm}
\usepackage{algorithmic}

\usepackage{newfloat}
\usepackage{listings}
\DeclareCaptionStyle{ruled}{labelfont=normalfont,labelsep=colon,strut=off} % DO NOT CHANGE THIS
\floatstyle{ruled}
\newfloat{listing}{tb}{lst}{}
\floatname{listing}{Listing}
\usepackage{comment}
\usepackage{adjustbox}
\usepackage{xcolor}
\newcommand\TODO[1]{\textcolor{red}{#1}}
\usepackage{caption}
\usepackage{subcaption}
\usepackage{enumitem}
\usepackage{multicol}
\usepackage{multirow}
\usepackage{adjustbox}
\usepackage{tcolorbox}
\usepackage{soul}
\usepackage{amsmath}
\title{Enhancing Content Moderation with Culturally-Aware Models}

\author {
    Alex J. Chan\textsuperscript{\rm 1, \rm 2},
    José Luis Redondo García\textsuperscript{\rm 1},
    Colm O'Donnell\textsuperscript{\rm 1}, \\
    Fabrizio Silvestri\textsuperscript{\rm 3},
    Konstantina Palla\textsuperscript{\rm 1}
}
\affiliations {
    \textsuperscript{\rm 1}Spotify, \textsuperscript{\rm 2}University of Cambridge, UK \textsuperscript{\rm 3}Sapienza University, Rome
}

\usepackage{bibentry}
\begin{document}
\nocopyright
\maketitle

\begin{abstract}
Content moderation on a global scale must navigate a complex array of local cultural distinctions, which can hinder effective enforcement. While global policies aim for consistency and broad applicability, they often miss the subtleties of regional language interpretation, cultural beliefs, and local legislation. This work introduces a flexible framework that enhances foundation language models with cultural knowledge. Our approach involves fine-tuning encoder-decoder models on media-diet data to capture cultural nuances, and applies a continued training regime to effectively integrate these models into a content moderation pipeline. We evaluate this framework in a case study of an online podcast platform with content spanning various regions. The results show that our culturally adapted models improve the accuracy of local violation detection and offer explanations that align more closely with regional cultural norms. Our findings reinforce the need for an adaptable content moderation approach that remains flexible in response to the diverse cultural landscapes it operates in and represents a step towards a more equitable and culturally sensitive framework for content moderation, demonstrating what is achievable in this domain.
\end{abstract}


\section{Introduction}
\label{sec:introduction}
Culture extends beyond geographical boundaries, fundamentally encompassing the collective beliefs and values that shape the behaviours and perception of a group of people. 
At a high level, it captures the many multifaceted aspects that influence human interactions and perspectives. 
Content moderation is tasked with preventing harmful or inappropriate content from proliferating, and here the tension between maintaining global online safety and respecting cultural nuances becomes particularly evident. Large-scale digital platforms often opt for standardised global rules, overlooking factors affecting content interpretation in diverse regions \cite{caplan:2018}. While intended to streamline content management, this one-size-fits-all approach inadvertently overlooks the importance of the cultural landscape. For instance, humour, which often relies on cultural references and context, can be particularly challenging to moderate \citep{jiang19}. A joke that is harmless in one culture may easily be misconstrued in another, leading to either potential censorship or, at the other end of the spectrum, the overlooking of genuinely inappropriate content. Similarly, handling content containing sexual references poses challenges due to differing regional boundaries on what is acceptable for general audiences.
Moving towards a culturally sensitive approach in content moderation is important in order to create a welcoming and respectful environment for users from diverse cultural backgrounds, while also maximizing the safely recommendable content. This concerns careful understanding and adaptation to different norms, values, delicate language nuances, and societal settings that characterise given communities when evaluating and managing content.

In this work, we show how integrating cultural awareness into modern moderation processes can significantly enhance content moderation through relatively simple fine-tuning. With the rise of powerful foundational language models, there is an opportunity to address the cultural nuances that are often overlooked in modern moderation methods. We begin by delineating culture as rooted within the geographical confines of a region and hypothesise that it is reflected in local media sources \citep{hanusch2016journalism, Hanush2017}. By training or fine-tuning language models on these culturally specific media sources, we can incorporate cultural insights and understanding into the moderation decision-making process. We make the following contributions:
(1) we introduce encoder-decoder models that are fine-tuned on media-diet data to inject cultural attunement;
(2) we propose a straightforward and extendable framework for content moderation that leverages these culturally attuned models through a three-step training process; and
%(2) we identify key areas within the existing content moderation pipeline where these culturally attuned models can be effectively integrated and 
(3) we illustrate, through a real case study of online podcast moderation, how culturally attuned models improve performance on local violative content detection and provide explanations that better align with the preferences of moderators familiar with local culture. 

%(3) we explore how these culturally-aware models can be integrated into a human-centric decision system, leveraging the advanced general reasoning capabilities of large language models.

\section{Current Moderation Approaches}
Content moderation remains one of the most challenging responsibilities for online platforms like X, Facebook, or YouTube. As they strive to prevent the proliferation of unsafe and untrustworthy sources, these platforms continuously invest in moderating violative content \cite{facebook, twitter, youtube}. 
Research indicates that automated tools are essential in addressing the pervasive issue of harmful online content. This necessity arises not only from the impracticality of relying solely on human moderation but also from the challenge of distrust in human-in-the-loop systems \cite{molina2022}. 
However, it is crucial to recognise that automated tools must be carefully implemented and overseen \cite{Christina2022} given that these ML systems often mirror the biases of their predominantly Euro-American creators, leading to the neglect of diverse cultural perspectives. For instance, \citet{Shahid:2023} shows that content moderation often worsens power imbalances, favoring Western-centric norms and marginalising expressions from users in the Global South.

The challenge becomes more pronounced when considering the cultural dimension of content moderation. \citet{Davani24} and \citet{victoria2014} stress that tasks like assessing deception — a behaviour that manifests variably across cultures — require a deep understanding of localised nuances. They emphasise the importance of tailoring ML systems to detect behaviours like deception by accounting for cultural specificities, a crucial step for Natural Language Processing (NLP) models to interpret and judge such behaviours accurately within various cultural contexts. Additionally, online reports \cite{yale2023} emphasise the importance of providing more nuanced insights into content moderation, particularly in highly diverse countries like India, rather than relying on broad statistical overviews. 
%Moreover, the intersectionality of gender and race further stratifies perceptions of moderated content, highlighting the intricate nature of content assessment from diverse identity perspectives \cite{hawkins2023}. \looseness=-1

To navigate these complex problems, culturally aware NLP systems are starting to be discussed and developed, with researchers arguing against the homogenisation of perspectives and advocating for a decolonisation of computational practices \cite{hershcovich2022, Birhane2020}. Such initiatives push towards a ``pluriversal epistemology'', embracing cultural diversity, and offer guidance on how to train NLP models and collect data that is relevant to achieve multiculturalism. However, cultural nuances are subtle, making the collection of culturally sensitive data, as well as the human-centred evaluation of these nuances, particularly challenging. %\citet{ringel2019} presents a transfer-learning framework that leverages bilingual corpora for classification tasks using no task-specific data, and evaluate its performance on formality classification and sarcasm detection tasks, showing that these approaches achieves comparable performance to task-specific methods directly trained on the two tasks. 
Examples include advances in detecting harmful content through multi-modal approaches \cite{pramanick2021, Sharma:2022}, and socially aware architectures such as in \citet{Yao2023}, where prompting strategies are used to integrate cultural knowledge into machine translation. These approaches demonstrate improvements over traditional systems when detecting harmful content or translating content, but despite these advancements, culturally aware NLP models remain an under-researched field. Recent efforts \cite{crosscult2024} are narrowing the gap, examining the alignment between pre-trained language models and cross-cultural values~\cite{arora2023}. \citet{akinade2023} demonstrate how language-specific tuning can refine machine translation architectures, while \citet{lee2023} expose the under-performance of hate speech classifiers when tested across different languages, pinpointing substantial degrees of cultural insensitivity. These studies underscore the relevance of research such as ours, advocating for enhanced models attuned to cultural subtleties. Finally, the assessment of cross-cultural alignment in models like OpenAI's ChatGPT suggests that while it aligns well within an American cultural context, it does not adapt as effectively to others\cite{cao2023}. The findings emphasise the significance of diversifying cultural perspectives in language technology.

%\section{Instilling Cultural Knowledge with Media Fine-Tuning}
\section{Culturally-Adaptive Moderation}
\label{sec:media-fine-tuning}

\begin{figure}
\centering
\includegraphics[width=.95\columnwidth]{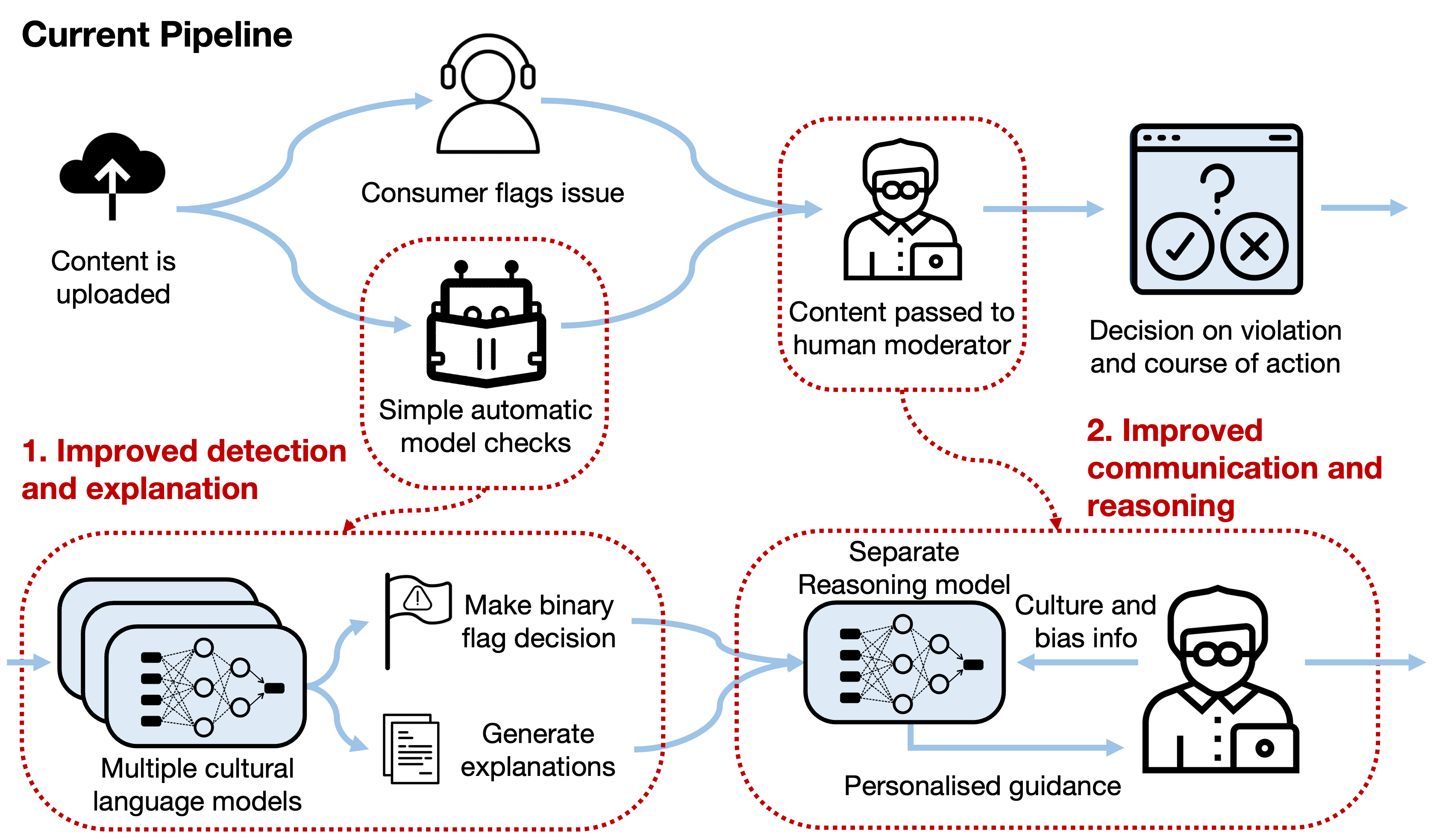}
\caption{\textbf{A culturally adaptive content moderation workflow}. Language models attuned to local culture 1.) enhance automatic detection and explanation of violations by disentangling local nuances and 2.) aid in aligning human annotators with global guidelines by serving as a reasoning engine, identifying and addressing mismatches in cultural norms between the moderator and location.}
\label{fig:adaptive_moderation}
%\vspace{-5mm}
\end{figure}

Most large-scale platforms use a content moderation pipeline, a simplified version of which is shown in the top half of Figure \ref{fig:adaptive_moderation}. Content like images and text is uploaded. Algorithms or users flag potential violations, which are then reviewed by human annotators trained to identify violations according to platform policies and a final decision is made. As discussed, this system follows global policies that often ignore cultural differences among users.

We propose enhancing the content moderation process by modifying the current system to better account for cultural differences during content evaluation. Our approach centres on using culturally adapted language models (LLMs) to improve decision-making. These models are customized to reflect specific cultural contexts and enhance two critical aspects of the moderation pipeline, shown at the bottom half of Figure  \ref{fig:adaptive_moderation}. First, they allow the system to make decisions that consider cultural nuances by providing both a numerical assessment of content violations and a natural language explanation that reflects how the content might be interpreted within a particular culture. This ensures that content is evaluated not only by global standards but also through the lens of local cultural sensitivities. Second, these culturally adapted models can assist human moderators by offering deeper insights into the cultural context of content. By integrating these models into a hybrid human-AI decision support system, we enable moderators to better understand the cultural implications of the content they are evaluating. This collaborative approach improves the accuracy and fairness of content moderation by ensuring that decisions are informed by both global and cultural perspectives.

%---
%The bottom half of Figure  \ref{fig:adaptive_moderation} shows how we can improve this process by using culturally adapted language models while minimally disrupting the overall workflow. First, we consider the automatic detection systems currently trained on global violations. Our approach adapts the pipeline by incorporating multiple culturally adapted language models, each tailored to a specific culture. These models provide both a numerical prediction regarding the content's violation status and a natural language explanation, reflecting how the content may be perceived within the local culture. If these models effectively predict human evaluations, they could serve as digital representations of local cultural sentiment. Second, using this array of cultural models can help human moderators better understand content in its cultural context. By using these models, or a relevant subset, the system can assist moderators in grasping the broader implications of content within various cultures.

\subsection{Media-diet Encoder-Decoders}
\begin{figure*}
\centering 
%\hspace{-12mm}
\includegraphics[width=.8\textwidth]{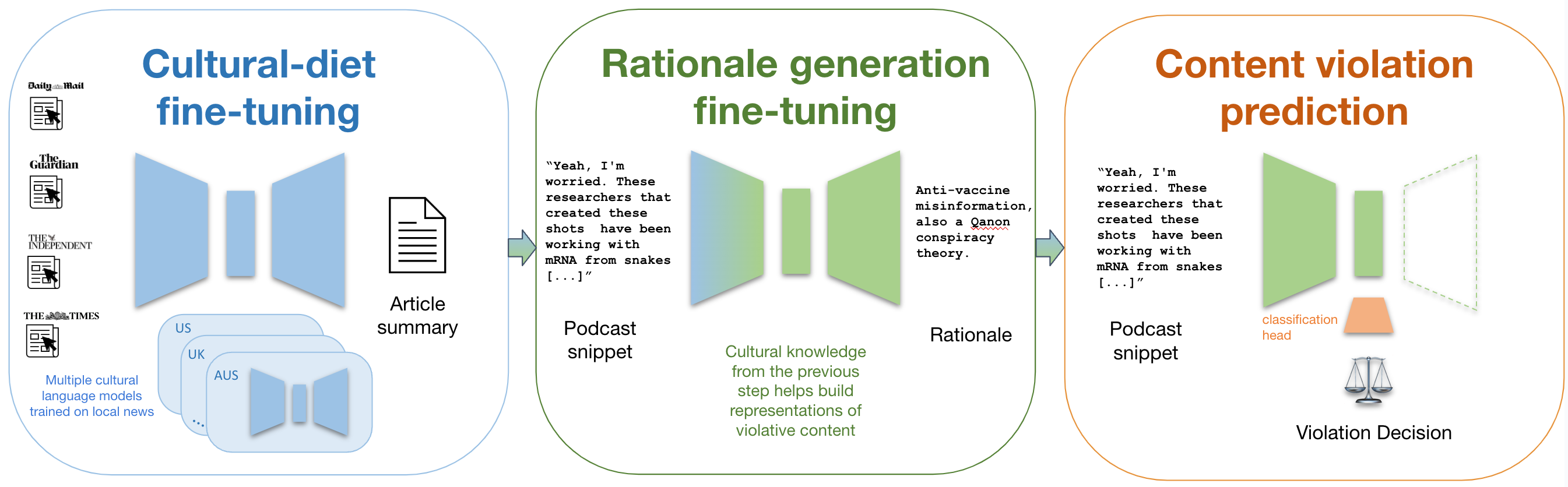}
%\vspace{-2mm}
\caption{\label{fig:methodology} \textbf{Proposed Three-Step Framework for Cultural Media-diet Violative Detection Models.} First, an encoder-decoder model is fine-tuned to summarise local popular news articles. Second, the pre-trained model is fine-tuned on generating moderator-written rationales of moderation decisions. Last, utilising the frozen encoder, a new classification head is trained to predict whether the content would be classified as a violation or not.` \looseness=-1}
\end{figure*}

At the core of our proposed modification to the moderation pipeline are the culturally adapted encoder-decoder architectures. We fine-tune these models to be sensitive to cultural nuances but also support content moderation tasks through a multi-stage training process; cultural-diet fine-tuning, followed by further refinement for generating explanations and identifying violative content. Figure \ref{fig:methodology} shows the proposed three-step training process.

During \textbf{cultural-diet tuning}, we employ a pre-trained language model and imbue it with a familiarity of the local culture. More specifically, we fine-tune an encoder-decoder language model on a summarisation task using \{\textit{article, summary}\} pairs sourced from a diverse range of local news outlets specific to each region resulting in media-diet models. We assume that fine-tuning the language model on news data collection from different regions results in distinct culturally-attuned models. In the second and third step, we further refine the culturally-attuned language model, focusing on two distinct tasks: (i) providing a rationale for the decision, and (ii) identifying content that violates established guidelines (i.e. a classification task). To \textbf{generate explanations}, we fine-tune the media-diet encoder-decoder of the first step on a sequence-to-sequence task of \{\textit{content, rationale}\} pairs. The encoder is then employed with the weights fixed as a component of a supervised task in which a classification head is fine-tuned using \{\textit{content, violation label (0/1)}\} pairs for the purpose of \textbf{detecting content violations} (last stage). %The pipeline uses media-diet at its core to ensure that the model is attuned to the local cultural context. This foundational step is crucial for tailoring the model's explanations and detection capabilities to the specific nuances of regional and cultural norms
The model undergoes cultural adaptation by fine-tuning on region-specific news. Media news was chosen because it offers a rich and timely source of region-specific information, reflecting the socio-political and cultural dynamics that shape local perspectives. 
In the next step, this cultural understanding is used to explain violations from a cultural perspective, ensuring contextually relevant justifications. Finally, the model applies this cultural insight to decide on content violations, with the decisions and scores reflecting its cultural adaptation. Each step of the pipeline builds on the media-diet, providing a continued training.

\section{Case Study: Global Podcast Moderation}
To showcase the effectiveness of our framework, we conducted a case study focused on detecting and explaining violative content on a global podcast platform.
We collated datasets for the following ten regions, which we consider as associated with distinct cultural contexts: United Kingdom, Canada, Australia, Nigeria, Malaysia, South Africa, Hong Kong, Kenya and India. 

\subsection{Culture-diet fine-tuning}
\paragraph{Dataset}
We constructed ten distinct media-diet datasets, each representing the specific culture, to support the cultural pre-training of our language models. Using the newapi.ai Python API \cite{newsapi}, we scraped online news articles from the top 50 sources in each of the ten cultures, ranked by local popularity. We curated articles published within the last 30 days and prioritized them based on their social media prevalence, retaining the top 50,000 articles or fewer if the total was lower. To allow for sequence-to-sequence training, we created summary targets using the pre-trained summarizing model \texttt{facebook/bart-summarize-cnn} \cite{lewis2020bart,wolf-etal-2020-transformers} to each article, resulting in \{\textit{article, summary}\} pairs. This approach ensures datasets are large, culturally diverse, and socially relevant, focusing on current trends rather than traditional writing quality. More details on the data collection can be found in the Appendix. %\ref{sec:appendix_methods}.

\paragraph{Model architecture and training}
We start with a set of models, one for each culture we consider, all using the same encoder-decoder transformer architecture: each model is initialized with a pre-trained BERT encoder and decoder \cite{devlin-etal-2019-bert}, alongside an additional randomly initialised binary classification head acting directly on the output of the encoder model. We intentionally selected the BERT model as our base due to its training on Wikipedia\footnote{https://www.wikipedia.org/} and BookCorpus \citep{Zhu_2015_ICCV}, which offer a broad knowledge base while minimizing bias by avoiding other online data sources. %However, biases can still be present due to the nature of the internet. 
Our objective was to minimise any pre-existing cultural bias within the model. This approach would provide greater assurance that any variations in performance downstream could be attributed primarily to cultural training. 
To adapt the base models to cultural norms, we train them on a sequence-to-sequence summarisation task using the \{\textit{article, summary}\} pairs: each model receives a media article and generates an informative summary. We selected summarisation over other pre-training tasks, like masked language modeling \cite{devlin-etal-2019-bert}, because it trains both the encoder and decoder and more closely aligns with generating explanations.

\subsection{Content Violation and Rationale Generation}

\paragraph{Dataset}
%Given access to a collection of local moderators from a popular large-scale podcast platform, we assembled a dataset comprising annotated podcast transcript snippets with binary labels indicating potential violations, along with text explanations clarifying reasons for violation, annotated by the moderators. 
We gathered a dataset with the help of local moderators from a popular large-scale podcast platform. The dataset includes podcast transcript snippets labelled as either violating or not violating, along with explanations from the moderators about why each snippet is considered a violation.
The resulting dataset includes $2,822$ positive violation examples and $4,393$ negative examples of \{\textit{snippet, label, rationale}\}, sourced predominantly from four markets: United States, United Kingdom, Australia, and Canada. For a detailed description of the data collection, curation and distribution see Appendix.%\ref{sec:appendix_methods}.

\paragraph{Model architecture and training}
We further fine-tune the entire encoder-decoder of the culturally trained models of the previous step on the task of generating explanations of violations (See Figure \ref{fig:methodology}, second stage). This is a supervised sequence-to-sequence task and we deploy the \{\textit{snippet}, \textit{rationale}\} pairs. 
Our fine-tuning process consisted of two distinct stages. In the first stage, each culturally attuned model was fine-tuned exclusively with violative examples specific to its culture. This step reinforced the model’s understanding of cultural nuances and maintained relevance to the cultural norms and sensitivities it was trained on. In the second stage, we expanded the fine-tuning to include all available rationales, regardless of cultural context. This broader fine-tuning was necessary due to the limited availability of culturally specific data and aimed to ensure that the models could generate coherent and contextually appropriate explanations even in the absence of extensive cultural data. This sequential approach allowed us to preserve cultural integrity while enhancing the models' robustness across diverse contexts. 

Following the fine-tuning stage on generating rationales, we use the embeddings from the fine-tuned encoder as input to a classification head. Each culture-specific classification head is further fine-tuned to determine whether the input is violative using the \{\textit{snippet, label}\} pairs.
%In our experimental setup, this process comprises two distinct procedures: initially, a stratified fine-tuning process focuses on providing each culturally attuned model from stage one solely with violative training examples (i.e., rationales) relevant to its respective culture. Subsequently, another process involves fine-tuning each culturally attuned model from stage one with all available rationales.

\section{Results}
\subsection{An Instilled Sense of Local Culture}
To assess the reliability of cultural-diet training (Figure \ref{fig:adaptive_moderation}, stage 1) and evaluate how well each model is attuned to its specific cultural context, we test their performance in summarizing media texts from various regions. We assume that models trained with different cultural contexts will produce distinct summaries of the same text, with summaries from models closely aligned to the cultural origin of the text being more accurate.
We use the ROUGE-1 metric \citep{lin:2004}, which measures lexical similarity between the generated summaries and reference summaries and serves as a proxy for understanding how closely each model's summaries reflect its cultural context. We examine the variability in the estimates within and across different corpora. Summarisation, while not the only possible approach for evaluating cultural attunement, offers a practical starting point. By analyzing how models handle summarisation of texts from different regions or cultures, we expect that models more attuned to the culture or region of the text will produce summaries that more closely match the reference summaries from that corresponding region.

\begin{figure*}
\centering
\begin{minipage}[b]{.45\textwidth}
  \centering
  \includegraphics[width=.8\textwidth]{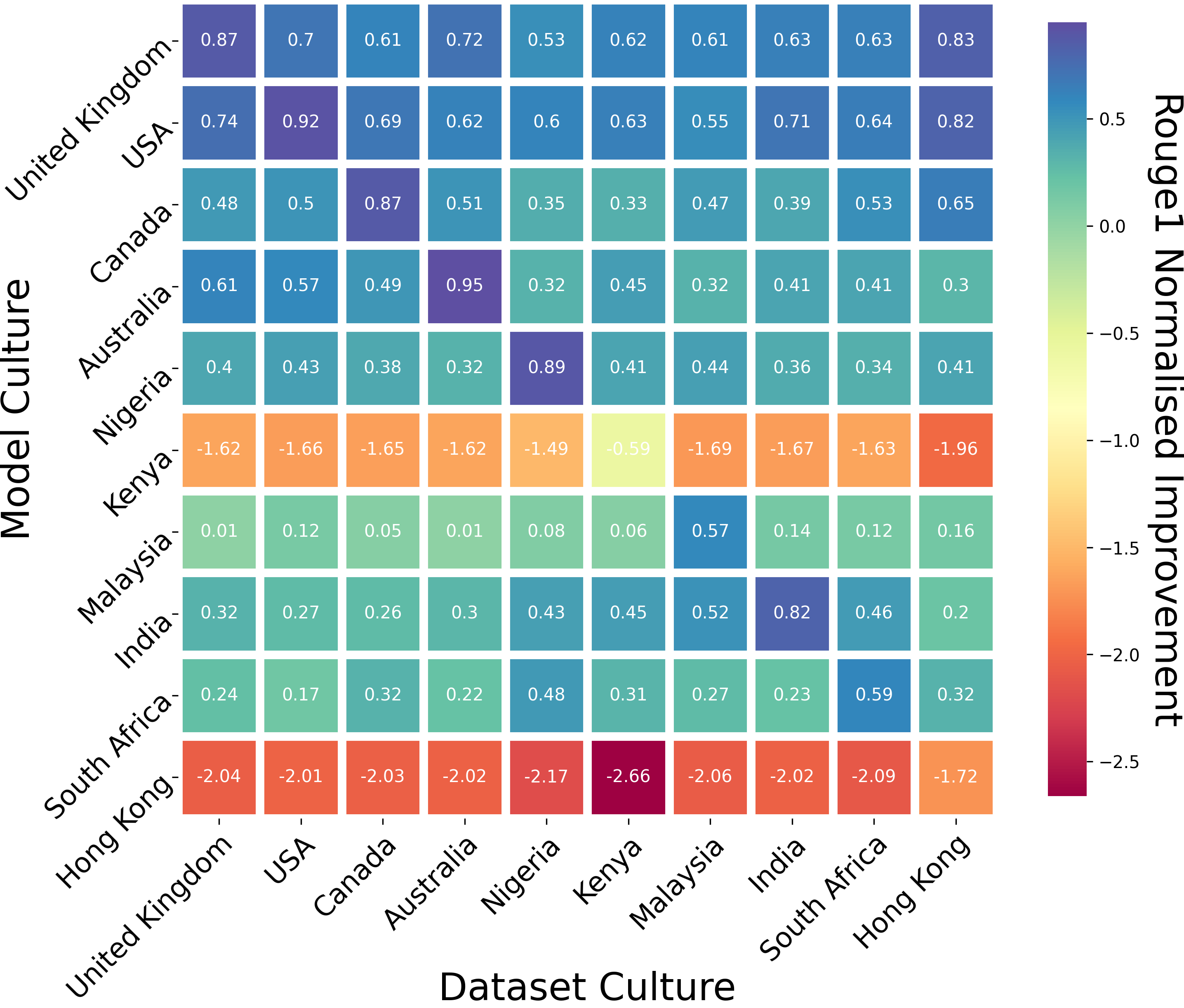}
  %\subcaption{\textbf{Media-diet Model Summarisation Performance.} A heatmap of the normalised improvement of the various cultural media-diet models on the test-sets of all cultures media-diet datasets. A strong leading diagonal indicates each model making proportionally larger gains in their own culture due to their training.}
  \label{fig:rouge}
\end{minipage}
\begin{minipage}[b]{.45\textwidth}
  \centering
  \includegraphics[width=.95\textwidth]{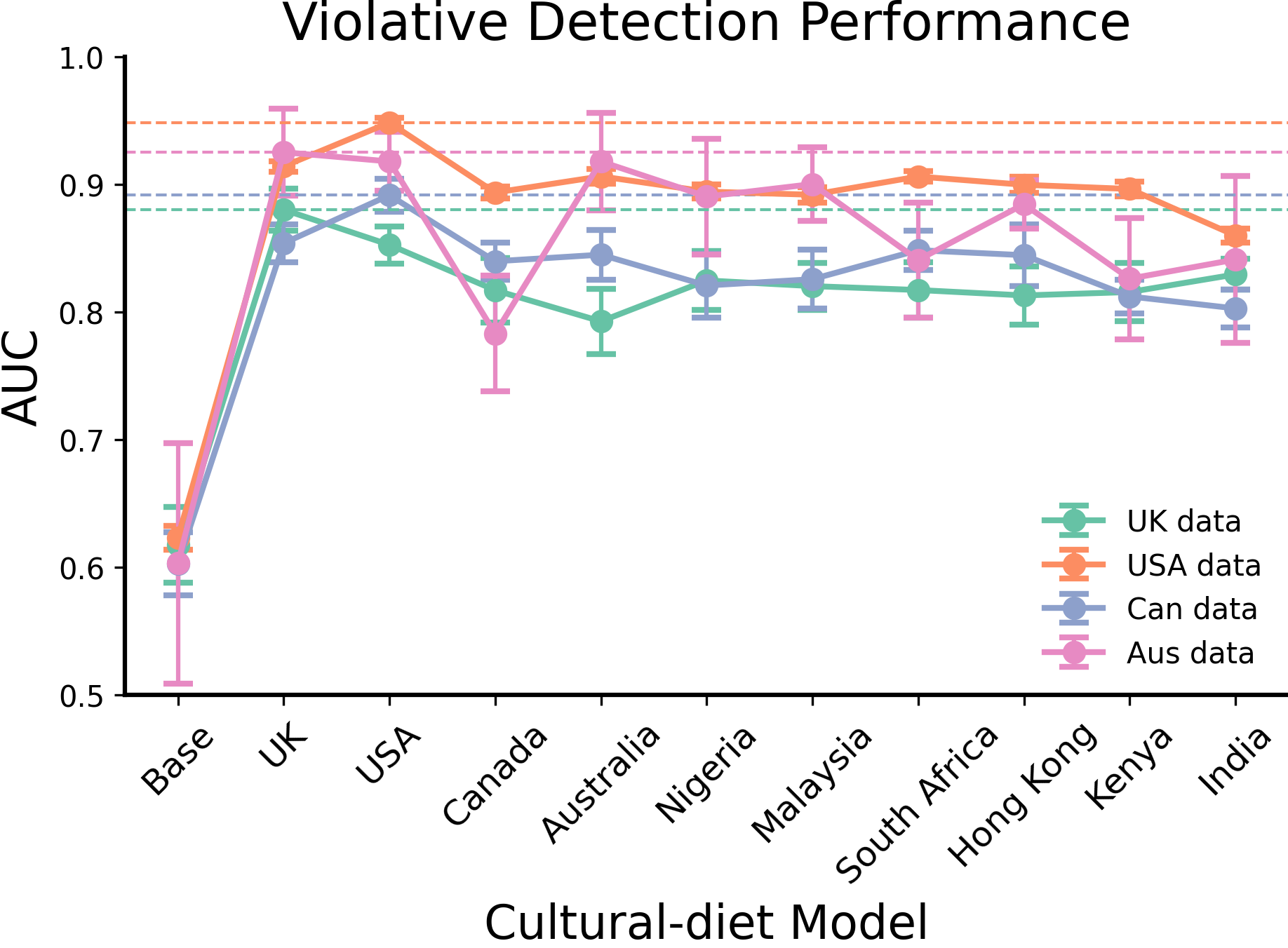}
  %\subcaption{\textbf{Violative Detection Performance}. We plot the performance of the cultural models over annotated content in local geographies as measured by AUROC. Test performance is split over content for four different locales.}
  \label{fig:perf_geo}
\end{minipage}
  \caption{Media-diet Model Performance. \textbf{(Left):} Heatmap showing normalised improvement of cultural media-diet models across test sets. A strong leading diagonal indicates each model making proportionally larger gains in their own culture. \textbf{(Right):} AUROC performance of the ten cultural-diet models and baseline on content from UK, USA, Canada and Australia.}
\label{fig:figures}
\end{figure*}

The heat-map in Figure \ref{fig:figures} (left) illustrates the ROUGE-1 normalized improvement across models trained on different cultural datasets (rows) and their performance on various cultural validation sets (columns). Each cell represents how well a model trained on a specific culture's data performs on texts from another culture. The diagonal values, which represent a model's performance on datasets from its own culture, are generally the highest, indicating that models are most effective at summarizing content from their training culture. Each model serving as an expert within its unique cultural domain, ideally owing to the culturally specific training data. Each model to excel within its own cultural context, while still being outperformed by other cultural models when applied to different datasets. This ensures that no single model dominates the others, with each model serving as an expert within its unique cultural domain. This suggests that the models are well-attuned to their respective cultural contexts. Exceptions are noted for the Kenya and Hong Kong models, where the diagonal values are not the highest, suggesting potential under-performance due to insufficient training data (see Appendix, Figure \ref{fig:article_count}). These deviations highlight areas where further training or data collection may be necessary to improve cultural attunement.

\subsection{Locally Improved Violation Detection}
\textit{How does cultural attunement impact model's ability to detect culturally sensitive content violations?} We evaluate all ten cultural models in the classification task, using data from four of the original ten cultures: the United States, United Kingdom, Canada, and Australia. These cultures were chosen because they provide a substantial number of violative examples, allowing for an accurate assessment of each model's performance. Our main goal is to understand how model performance varies across these target cultures, influenced by the culture-diet training on media data. \looseness=-1
%Performance is evaluated using the Area Under the Receiver Operating Characteristic Curve \citep[AUROC]{hanley1982meaning}. 

Figure \ref{fig:figures} (right) shows the performance of each cultural model at stage three in our pipeline (Figure \ref{fig:methodology} and a base model without media-diet pre-training across the four test cultures. The base model (BERT) representations are kept frozen with only the classification head fine-tuned on \{\emph{snippet, label}\} pairs from all four regions. The highest performance achieved by any model is marked on the y-axis with dotted lines. 
All cultural models outperform the base model, an improvement that may be attributed to the additional training with more data; irrespective of cultural differences, fine-tuning the model with media data ensures its knowledge remains current, which is essential for identifying violations associated with contemporary context. Our choice of the base model, with outdated training data, further amplifies this effect. This underscores the potential of keeping the model up to date through media-based training for effective violation detection, even if highly specific cultural dataset is not available.

Training on media from culturally distinct regions, such as Indian media, yields a significant improvement in the base model's ability to classify violations originating from the UK market. This phenomenon could be attributed, in part, to the historical legacy of colonisation. The common use of English language media in both India and the UK has likely contributed to the models' ability to bridge cultural and linguistic gaps between these regions \cite{crystal2007english}. This historical connection underscores the impact of cultural and linguistic influences on machine learning models, even across geographically distant regions. We can see similar patterns to the previous evaluation of instilled culture on the summarisation task; the models that achieve the highest AUROC on data from a particular market generally are the ones that have received the cultural training from the same market.

Both the UK (AUC=0.880, $95\%$CI:$(0.864,0.897)$) and the US (AUC=0.948, $95\%$CI:$(0.945,0.952)$) models achieve statistically-significant highest scores (UK: t(17.7)=2.503, $p=0.022$; USA: t(17.9)=12.286, $p<0.01$) among all other models on their respective cultural datasets. The Australian model is among the highest performing for the Australian data, but there is no one statistically-significant highest performer (UK AUC=0.925, $95\%$CI:$(0.891,0.959)$, AUS AUC=0.918, $95\%$CI:$(0.880,0.956)$, t(17.8)=0.283, $p=0.781$). We observe much wider confidence intervals for the Australian evaluation, in part due to a much smaller set of positive violative examples (see Appendix, Figure \ref{fig:violative_categories}).
The Canadian model also performs competitively, and is only significantly outperformed by the US model (CAN AUC=0.839, $95\%$CI:$(0.825,0.854)$, USA AUC=0.891, $95\%$CI:$(0.878,0.904)$, t(17.8)=5.341, $p<0.01$). This may be due to the higher number of violative instances in the US dataset (see Appendix, Figure \ref{fig:violative_categories}), allowing the US model to capitalise on those training data points. Additionally, the cultural proximity between the US and UK, the most closely related pair of markets examined, might diminish the impact of cultural training, as their similarities could reduce the necessity for highly specialized models.

Our dataset is limited in moderator-labelled data from regions outside the US, UK, Australia, and Canada, which are often categorized as ``\emph{western}'' markets due to shared, cultural, economic and political traits. This limitation makes it challenging to fully assess the impact of cultural training. However, models trained on non-western data perform as expected, showing lower performance in western markets.

\begin{comment}
   \TODO{Impact of cultural training on identifying content violation
\begin{itemize}
    \item Performance of cultural models across geographically stratified content (content moderation task). Example Figure \ref{fig:perf_geo}.
    \item Show correlations between model scores and content. Barplots (with error bars) of the scores of each cultural model on the set of positive cases across geographies. E.g. sns.barplot(df[df[‘violation’]==1][‘score’]). This plotting might help us see whether the confidence of the model is stronger for local cases.
    \item For each local data and for each policy type: plot barplots of the AUC of each cultural model.
    Can we see policy categories where a cultural model performs better? This plot might provide insight to whether the cultural bias is more evident in specific policy cases.
\end{itemize}
} 
\end{comment}

\paragraph{Exploring the model's predictions}

\begin{figure}
\centering
\begin{subfigure}[b]{\columnwidth}
  \centering
  \includegraphics[width=.95\columnwidth]{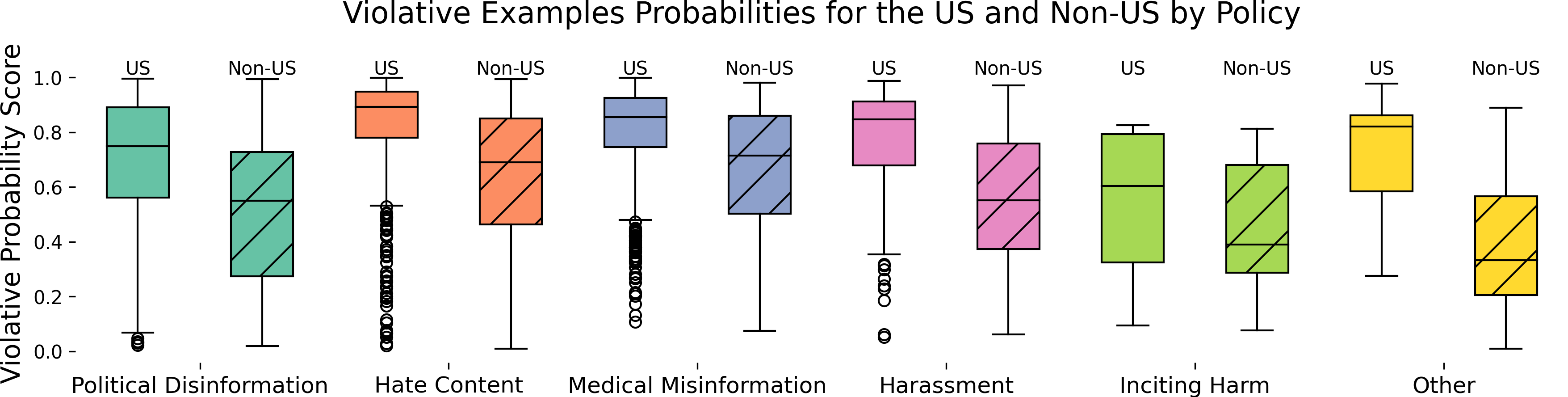}
  \caption*{}
  \label{fig:boxplot_usa}
\end{subfigure}% \\
%\vspace{0.1in}
\\
\begin{subfigure}[b]{\columnwidth}
  \centering
  \includegraphics[width=.95\columnwidth]{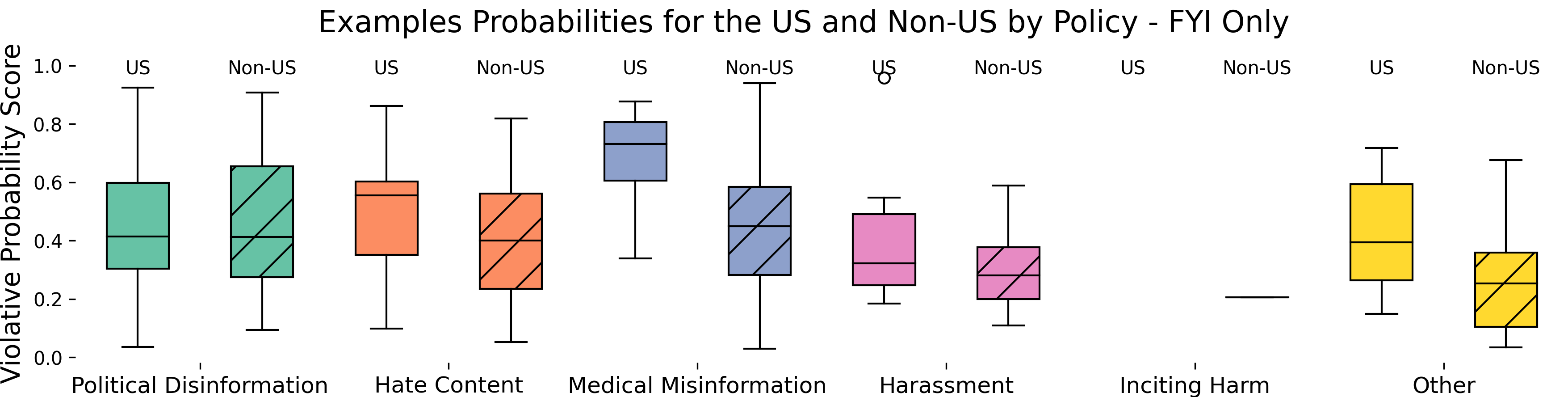}
  \caption*{}
  \label{fig:boxplot_usa_fyi}
\end{subfigure}
\caption{\textbf{Exploring the United States Model.} Score distribution for (top) violative cases and (bottom) `FYI' subset by the US model stratified to US and non-US origin cases.}
\label{fig:usa_policy}
\end{figure}
We focus on the US model to gain a better understanding of the reasons behind performance gains, as it was trained on the largest dataset, providing clearer signals (see Appendix, Figure \ref{fig:violative_categories}).  In Figure \ref{fig:usa_policy} (top) we examine the probability assigned by the US model to content labelled as violative, considering both US-origin content, i.e. content identified as violative by local annotators in the US, and content from other regions. The model consistently assigns higher scores to US-origin content across all policy types, indicating greater confidence in its predictions, with scores closer to 1. This higher confidence is further supported by Figure \ref{fig:figures} (right), where the model accurately classifies more examples. We observe significant distributional shifts when examining variations by violation type, particularly in the \emph{``Hate Content''} category likely due to cultural factors that influence how hate is expressed and perceived in different societies.
In contrast, the \emph{``Inciting Harm''} category shows minimal variation between US and non-US cases, reflecting the straightforward nature of this violation with less room for cultural interpretation. The the \emph{``Medical Misinformation''} category also shows substantial shifts, influenced by geographically variable issues like COVID-19 conspiracy theories, which have become politicized in distinct regions \cite{romer2020conspiracy}.

We evaluate the models on the ``FYI leads'' dataset,  which contains content flagged by local moderators as non-violative globally but potentially concerning within a local context. The term ``FYI'' stands for ``For Your Information'', indicating that this dataset is intended to alert decision-makers to emerging risks and prompt guideline updates if necessary. Evaluating the models on these cases allows us to assess their ability to recognize cultural nuances beyond strict violations.
In the lower part of Figure \ref{fig:usa_policy}, we observe that FYI cases receive lower scores compared to clear-cut violations (shown in the upper part).
For most of the policy categories, there is not significant shift in the score distribution. However, we notice a discernible shift in the score distribution between US and non-US examples in the `Medical Misinformation' and `Hate Content' categories. These shifts likely stem from the interplay between cultural sensitivity and local understanding. Content related to medical and hate issues is often highly sensitive to cultural nuances, and models that are attuned to these local contexts, much like local moderators, may be better at identifying potentially problematic content, even if it does not strictly violate global guidelines. This local perspective can result in higher scores for such content.

\subsection{Culturally Aligned Model Explanations}
\begin{figure}
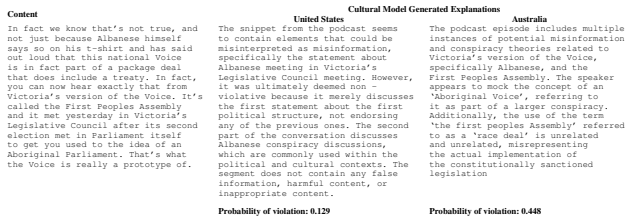

  \centering
  \begin{adjustbox}{max width=\linewidth}
  \begin{tabular}{p{8cm}p{8cm}p{8cm}}
    %\hline
    \multirow{2}{*}{\textbf{Content}} &  \multicolumn{2}{c}{\textbf{Cultural Model Generated Explanations}} \\
     & \multicolumn{1}{c}{\textbf{United States}} & \multicolumn{1}{c}{\textbf{Australia}} \\
    %\hline
    \texttt{In fact we know that's not true, and not just because Albanese himself says so on his t-shirt and has said out loud that this national Voice is in fact part of a package deal that does include a treaty. In fact, you can now hear exactly that from Victoria's version of the Voice. It's called the First Peoples Assembly and it met yesterday in Victoria's Legislative Council after its second election met in Parliament itself to get you used to the idea of an Aboriginal Parliament. That's what the Voice is really a prototype of.} & \texttt{The snippet from the podcast seems to contain elements that could be misinterpreted as misinformation, specifically the statement about Albanese meeting in Victoria's Legislative Council meeting. However, it was ultimately deemed non - violative because it merely discusses the first statement about the first political structure, not endorsing any of the previous ones. The second part of the conversation discusses Albanese conspiracy discussions, which are commonly used within the political and cultural contexts. The segment does not contain any false information, harmful content, or inappropriate content.} & \texttt{The podcast episode includes multiple instances of potential misinformation and conspiracy theories related to Victoria's version of the Voice, specifically Albanese, and the First Peoples Assembly. The speaker appears to mock the concept of an `Aboriginal Voice', referring to it as part of a larger conspiracy. Additionally, the use of the term `the first peoples Assembly' referred to as a `race deal' is unrelated and unrelated, misrepresenting the actual implementation of the constitutionally sanctioned legislation} \\ 
     & \vspace{1mm} \textbf{Probability of violation: 0.129} & \vspace{1mm} \textbf{Probability of violation: 0.448} \\
    %\hline
  \end{tabular}
  \end{adjustbox}
  \caption{\textbf{Cultural Model Explanations}. An example of how different cultural models explain an ``FYI'' lead created in the Australian market.}
  \label{fig:cultural_explanations}
\end{figure}
\textit{How does cultural adaptation affect the relevant explanations for content violation decisions?}
Our goal is to determine whether adapting a model to specific cultural contexts improves its ability to provide culturally aligned \emph{explanations}. To assess this, we fine-tune the cultural models to generate explanations similar to those provided by moderators when flagging content.

Figure \ref{fig:cultural_explanations} illustrates this with an example from an Australian podcast discussing the "Voice", a proposal for Indigenous representation in Australia’s Constitution. This topic involves issues of form, symbolism, political dynamics, and diverse Indigenous perspectives. Moderators classify it as FYI-lead, suggesting it should be monitored for future concerns rather than flagged as a violation. The US model deems it irrelevant, causing uncertainty about potential violations. The Australian model assigns a higher probability of violation, acknowledging concerns like the term "Aboriginal Voice" and misinformation risks. It demonstrates a superior grasp of the content's cultural context, correctly identifying potential violations and addressing concerns such as the use of terminology and misinformation risks. 

\begin{figure}
\centering
\begin{subfigure}{\columnwidth}
  \centering
  \includegraphics[width=\columnwidth, height=2cm]{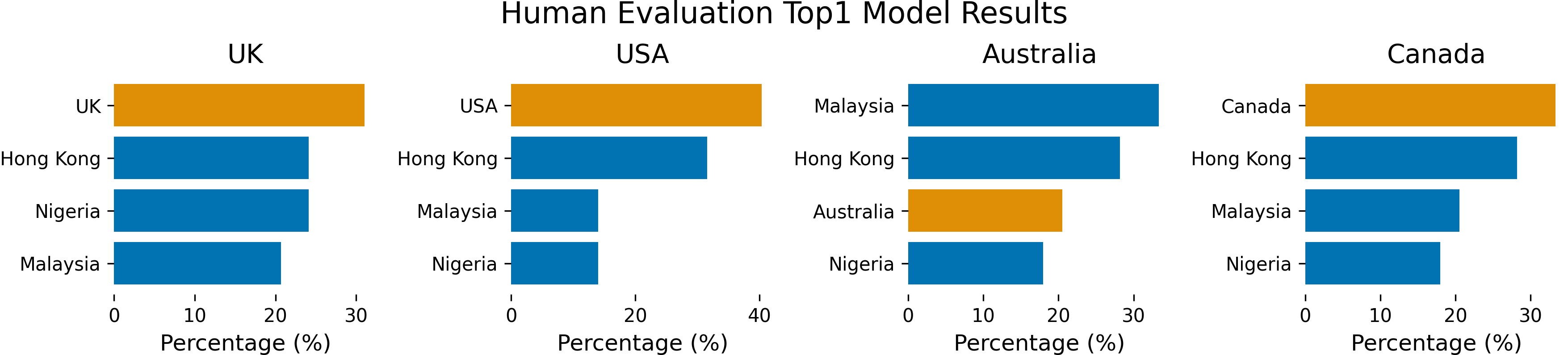}
  \caption*{}
  %\label{fig:human_eval_res}
\end{subfigure}%
\\
%q\vspace{0.2in}
\begin{subfigure}{\columnwidth}
  \centering
  \includegraphics[width=0.6\columnwidth]{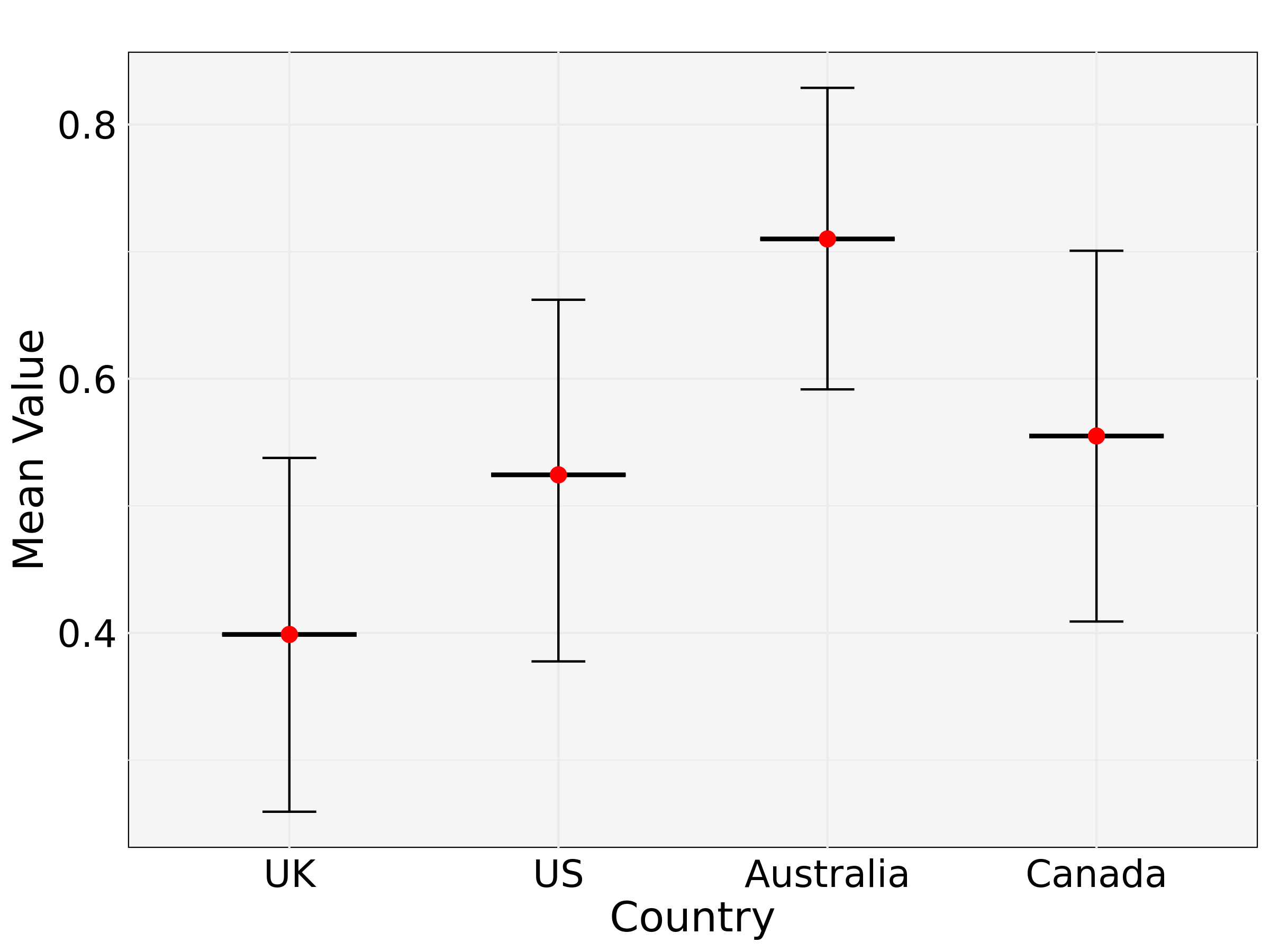}
  \caption*{}
 %\label{fig:kendall}
\end{subfigure}
\caption{\textbf{Human Evaluation.} Top: Percentage of examples where local experts selected each model as their first choice. Bottom: Kendall's coefficient of concordance with confidence intervals for each country.}

\label{fig:annotators}
\end{figure}

Evaluating cultural differences in model explanations is challenging. Although frameworks like Hofstede's \citep{Hofstede1980} can analyze cultural norms in text, they may miss subtle insights that only people from those cultures can provide. We involved human annotators from each of the considered locales: three from the UK, three from the US, two from Canada, and two from Australia. Each annotator reviewed explanations generated by four different models, including one model trained on data from their own culture. For each of twenty different content pieces, they ranked the four explanations in order of preference. For three out of the four models, i.e. UK, US and Canada, the aligned model was the one most often selected as the first choice of the expert moderators, as shown in Figure \ref{fig:annotators} (top).

To measure agreement among raters, we use Kendall's Tau coefficient of concordance ($W$) \citep{kendall1938measure}, where values range from -1 (strong disagreement) to 1 (strong agreement). The Kendall's $W$ values vary across countries (Figure \ref{fig:annotators}, bottom). Australia shows high agreement ($W=0.71$, $95\%$CI:($0.59,0.83$)), the UK shows lower agreement ($W=0.4$, $95\%$CI:($0.26,0.54$)), and the US and Canada moderate ($W=0.52$, $95\%$CI:($0.38,0.66$) and $W=0.56$, $95\%$CI:($0.41,0.70$)). This suggests that Australian annotators found the explanations more culturally aligned compared to their UK counterparts. 
Expanding the analysis to include the cultural trio (Hong Kong, Malaysia, and Nigeria) used for non-local explanation, Australian annotators may perceive stronger cultural resemblances. Additionally, non-significant p-values for Kendall coefficients, i.e. Australia: $0.28$, UK: $0.31$, US: $0.13$ and Canada: $0.22$, suggest observed patterns may not significantly deviate from randomness, indicating limitations in detecting subtler effects possibly due to the study's sample size (see Appendix).

\section{Discussion}
Our results demonstrate that cultural adaptation in content moderation can be effectively achieved by fine-tuning language models with media diets. Traditional moderation approaches often overlook the depth of cultural context, relying on simpler, one-size-fits-all methods. Our approach leverages the nuanced understanding provided by media-diet attunement, enabling the model to generate culturally relevant explanations and accurately detect violations. In a case study on a global podcast platform, we validated our approach across diverse cultural contexts, showing its potential to significantly enhance content moderation systems. With further refinement, this method could complement existing moderation practices, offering a more tailored and equitable solution for different regions and communities. The media-diet, for example, could be supplemented by incorporating data from local social media platforms and traditional oral narratives to capture a broader cultural spectrum.

While our framework provides an effective approach to cultural adaptation, it also serves as a foundation that may inspire alternative architectures. Practitioners have the flexibility to explore different configurations within this framework to best suit their specific needs and preferences. For instance, one might choose to reverse the sequence of tasks, training for violation detection before rationale generation, or even integrate these tasks simultaneously.
%Tailoring content moderation systems to account for cultural differences is essential for addressing the limitations of one-size-fits-all global policies. Our study demonstrates the efficacy of adapting language models with cultural knowledge to enhance moderation across diverse regions. A case study on a global podcast platform reveals that incorporating culturally attuned models into the moderation pipeline results in improvements in both detection accuracy and explanation relevance. The cultural-diet fine-tuning process ensures that models better reflect local norms and sensitivities, leading to more precise identification of content violations and contextually appropriate explanations.

\subsection{Future Directions}
Looking ahead, we envision this framework evolving into a practical tool for real-time assistance to content moderators offering instant violation explanations and decisions with insights into local cultures.
To explore this potential, we have developed a prototype that demonstrates how this framework could work in practice and can be seen in Appendix. Recent advances in large causal language models have produced models that can effectively act as assistants and process novel information \cite{achiam2023gpt}. We propose the utilisation of such a model as a reasoning engine (Figure \ref{fig:methodology}, bottom right), which incorporates the cultural models as \emph{tools} \citep{schick2023toolformer} allowing it to invoke and process their output. Paired with a ReACT strategy \citep{yao2023react}, this framework enables the larger general model to function as a central reasoning engine. Moderators would have access to this engine, consulting it as needed. It can process information about the moderator using it, the consumer's culture and the content creator. Subsequently, it can invoke any appropriate cultural model before providing a concise summary of the findings to the moderator in an efficient manner. Examples of this system in use can be found in Appendix. Such a system further promises to reduce the manual workload for annotators but also to guide policy refinement by identifying distinctions that should be incorporated into policies for different cultural backgrounds \cite{openai_moderation}. However, it is vital to address biases introduced during training, including pre-training of language models and cultural considerations \citep{Schramowski2021LargePL}. Like any AI application, diligent oversight is crucial for verifying and improving results.

\subsection{Limitations}
\label{sec:limitations}
The core of our approach lies in fine-tuning language models to adapt them to cultural norms. While we initially focused on summarisation of online news articles, we recognise alternative methods like masked-language modelling \cite{devlin-etal-2019-bert}. Further efforts could be made to determine which language task could be more effective to infuse cultural-specific knowledge into the models. %Simultaneously, our approach is adaptable, allowing for various configurations and architectural adjustments.

Leveraging news media articles offers the advantage of data produced daily, capturing the fast-evolving and changing nature of cultural norms. Broadening our data collection to include a wider range of sources, such as community news websites, forums, and oral histories, would enrich the cultural training study. Furthermore, a more systematic analysis is needed to cover dimensions such as pre-existing bias in the language models used \citep{Schramowski2021LargePL}, bias in the new articles beyond cultural norms, or even norms that extend beyond current cultural understanding. 

The study adopts a country-centric view of culture, while recognizing that culture often transcends geographical boundaries. While this viewpoint may not fully capture intra-national cultural diversity, it aligns with established concepts in cultural and social sciences \cite{house2004}. Expanding this framework to include broader cultural communities requires extensive document collection and careful archival work, though systemic challenges in resourcing these efforts persist \citep{Wakimoto2013}.

We mainly focus on the English language and Western cultures due to limited moderator-labelled data from non-English and non-Western regions, hindering precise assessments of cultural training's global impact. Our framework can be extended to other languages by collecting suitable data. Moreover, our framework is presently constrained to the textual modality. Incorporating audio and visual modalities can enhance its capacity to capture cultural norms.%;researchers can seek collaborations in non-Western regions to gather more diverse data. 
%Expanding data collection efforts beyond Western countries will enhance our understanding of cultural influences on model performance in detecting violations.
% We primarily focus on Western cultures, potentially limiting the representation of global cultural diversity due to the scarcity of moderator-labelled data from regions beyond the US, UK, Australia, and Canada. Given the shared characteristics of these regions, assessing the precise impact of cultural training on global model performance is difficult. To overcome this limitation, researchers can seek collaborations in non-Western regions to gather more diverse data. Expanding data collection efforts beyond Western countries will enhance our understanding of cultural influences on model performance in detecting violations.

Finally, increasing the number of annotators per country and conducting more detailed surveys with additional questions could strengthen the insights from the alignment with annotators, possibly revealing more significant trends. Revisiting and refining the annotators study design, including question selection, and exploring deeper explanations may provide clearer insights into cultural dynamics.
% Considering the complexities of cultural attunement and perceptions among annotators from different countries, it might not be advisable to completely abandon further attempts at explanation or analysis with human annotators. To enhance future investigations, expanding the sample size beyond two/three annotators per country and conducting more comprehensive surveys (more questions) may unveil stronger, statistically significant trends. Reassessing the methodology, possibly refining the selection of questions, and exploring more complex explanations might offer deeper insights into the subtle dynamics between cultures.

\bibliography{aaai25}

\appendix
\section{Appendix}
\subsection{Methods}
\label{sec:appendix_methods}
The hardware setup used for the experiments comprised machine V100 with 32GB GPU memory. %All experiments were done using Python 3.9.16 and Pytorch version 1.13.

%\subsection{Methods}
%\label{sec:methods}
% All our models follow the same encoder-decoder transformer architecture, the base being formed of a pre-trained encoder and decoder both initialised as a BERT model \cite{kenton2019bert}, alongside an additional randomly initialised binary classification head acting directly on the output of the encoder model.
% We intentionally chose the BERT model as a base model to work on because it was trained on Wikipedia and BookCorpus, which aim to provide a broad knowledge base without incorporating any other mixture of online data. However, biases can still be present due to the nature of the internet. Our objective was to minimize any pre-existing cultural bias within the model. This approach would provide us with greater assurance that any variations in performance downstream could be attributed primarily to cultural training. An outline of the methodological training setup is shown in Figure \ref{fig:methodology}.

\subsubsection{Media-diet Cultural Fine-tuning}
\begin{figure*}[h]
\centering
\begin{subfigure}{.29\textwidth}
  \centering
  \includegraphics[width=\linewidth]{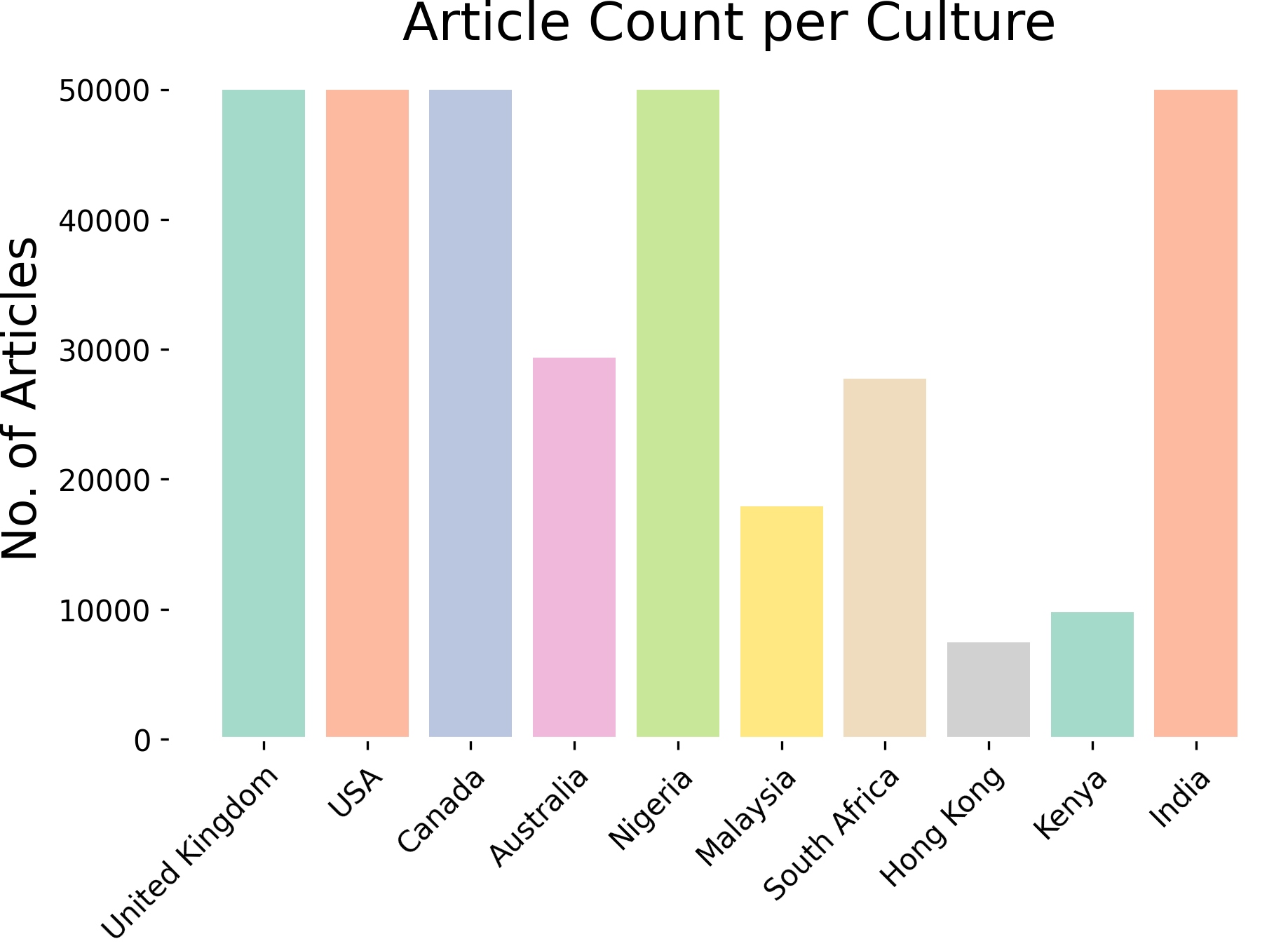}
  \caption{}
  \label{fig:article_count}
\end{subfigure}%
\begin{subfigure}{.29\textwidth}
  \centering
  \includegraphics[width=\linewidth]{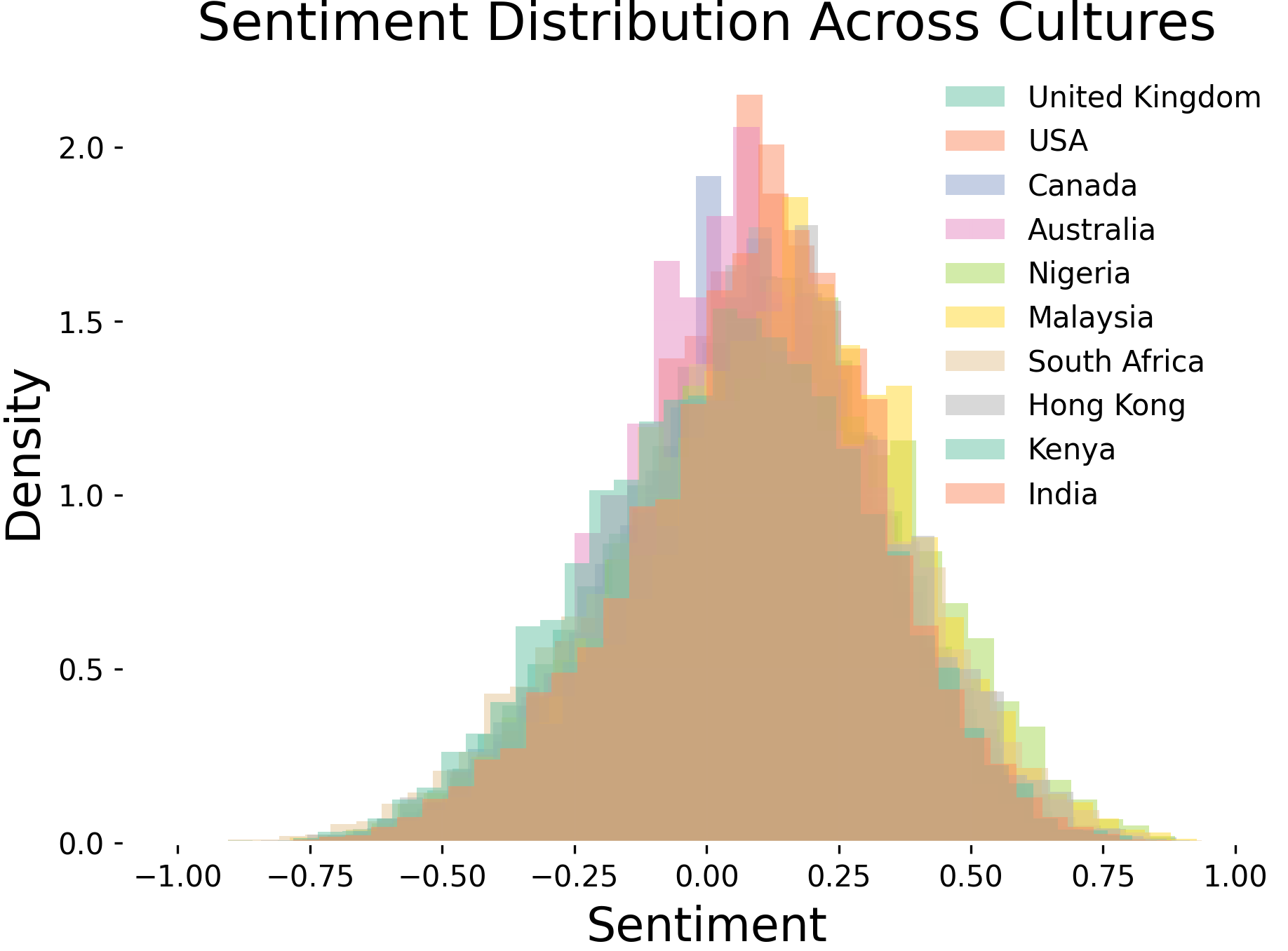}
  \caption{}
  \label{fig:sentiment}
\end{subfigure}
\begin{subfigure}{.39\textwidth}
  \centering
  \vspace{-5mm}
  \includegraphics[width=\linewidth]{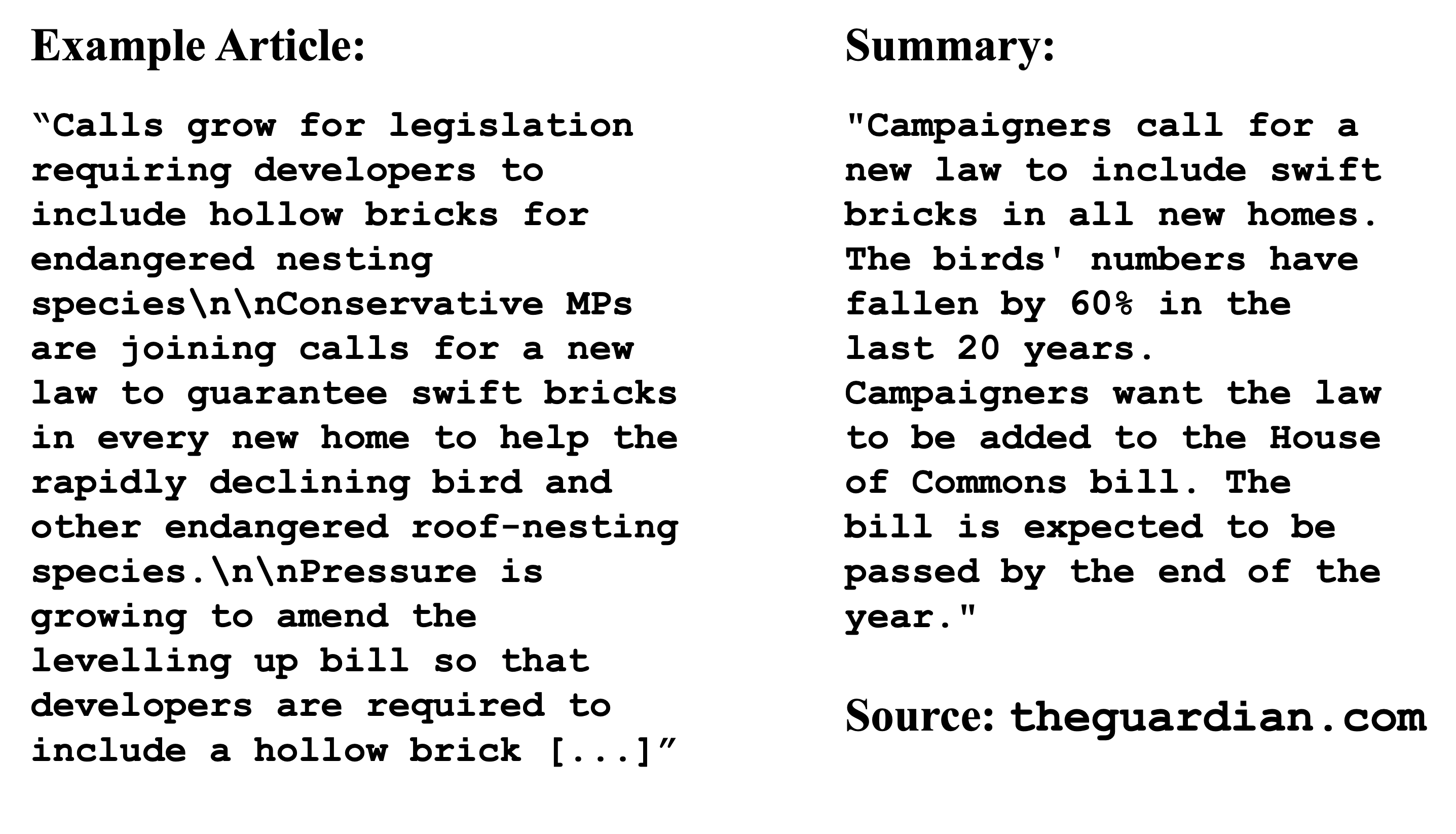}
  \caption{}
  \label{fig:data_example}
\end{subfigure}
\caption{\textbf{Understanding the Media-diet.} \textbf{(a)} The number of articles in each of the cultural datasets. 
\textbf{(b)} A histogram of the sentiment distribution per culture.
\textbf{(c)} Example article and generated summary from the UK dataset that is used for training.}
\label{fig:cultural_data}
\end{figure*}
\begin{figure}
\centering
\includegraphics[width=0.5\textwidth]{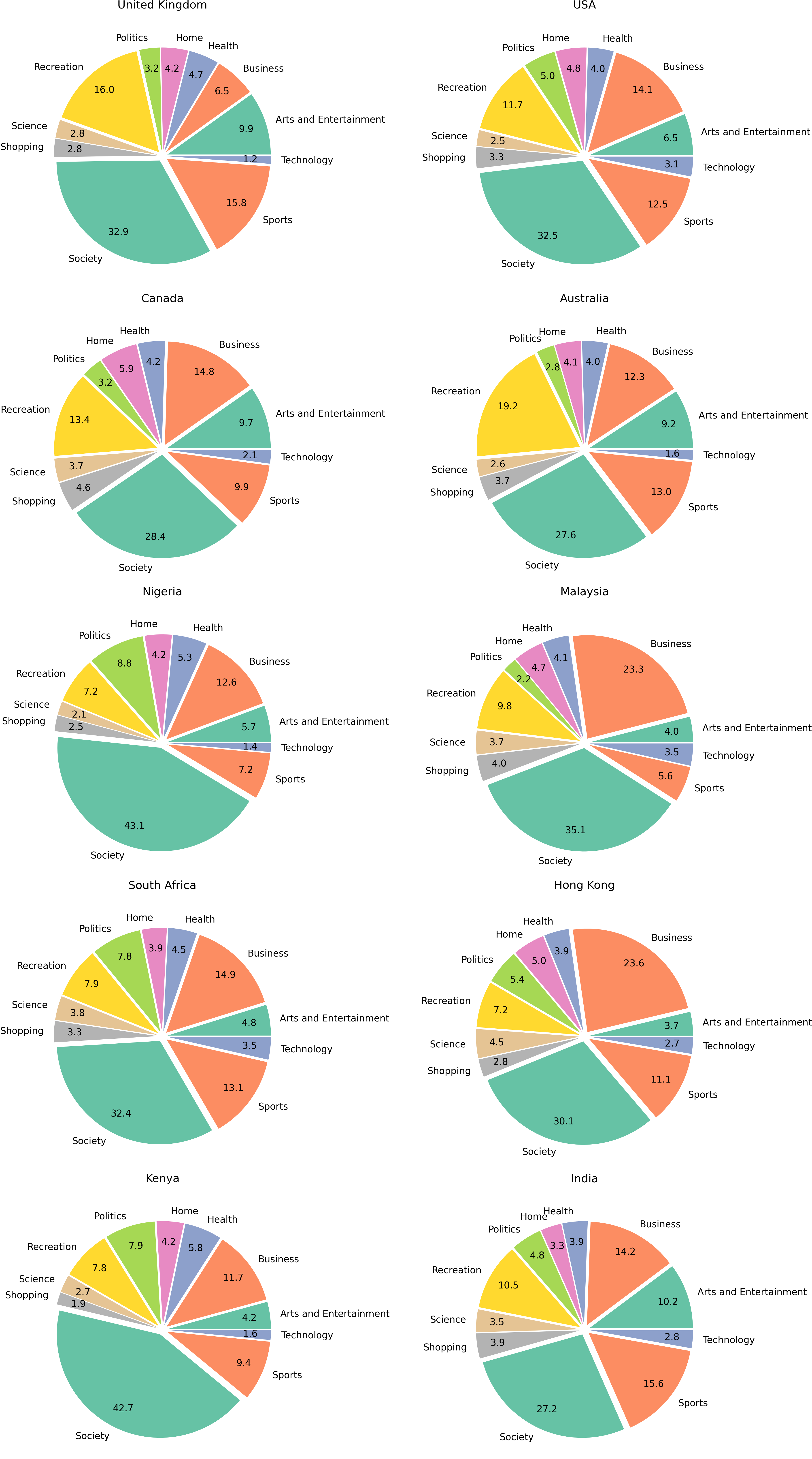}
\caption{\label{fig:categories} Cultural media-diet topic categories breakdown in percentage.}
\end{figure}

\paragraph{Dataset}
We constructed ten distinct media-diet datasets, each tailored to a specific culture. These datasets serve as a foundational resource for the cultural pre-training of our language models. We scraped online news articles using the newapi.ai \cite{newsapi} Python API. For each culture, we rank news sources within their respective geographic locations based on newapi.ai popularity, selecting the top 50 sources. We then curate articles from these sources published within the last 30 days and prioritize them based on their `social score', reflecting their prevalence on social media. The top 50,000 articles are retained, or all if the total falls below this threshold. Finally, we generate summary targets for our sequence-to-sequence task by applying a pre-trained summarizing language model, \texttt{facebook/bart-summarize-cnn} \cite{lewis2020bart,wolf-etal-2020-transformers} to each article. This comprehensive approach aims to construct large and well-balanced datasets sourced from diverse outlets, capturing the current cultural zeitgeist, with a distinct emphasis on social significance rather than conventional quality-of-writing metrics.
Figure \ref{fig:cultural_data} shows additional information about the collated datasets, including an example article and summary (Figure \ref{fig:data_example}). We are mainly interested in these plots to show that the different datasets are relatively balanced in terms of article sentiment (Figure \ref{fig:sentiment}, shows the normalised histograms of each culture overlap to a high degree) as well as the distribution of topics (See Figure \ref{fig:categories}). 

\paragraph{Model architecture and training}

In order for our models to learn culturally informed representations of the input text, we pre-train them on the \{\textit{article, summary}\} pairs constructed for each target country as described above. This is done in a supervised fashion on a sequence-to-sequence summarisation task; each model receives a media article and tasked with generating an informative summary of said article as the output.
There are multiple pre-training tasks we considered including alternatives such as masked language modelling. However, ultimately, we chose the summarisation because it offers the advantage of training both the encoder and decoder, also aligning closely with the objective of generating an explanation given an input.
All of the models are initialised with a pre-trained encoder and decoder both of which are based on the BERT model, forming a bert2bert architecture. Specifically, we used the \texttt{bert-large-cased} from the HuggingFace model repository \cite{wolf-etal-2020-transformers}. During encoding, the article passes through the network's encoder, resulting in a 1024-dimensional vector of real numbers for each token in the input. Subsequently, the decoder predicts the summary one token at a time, taking both the embedding and the previously generated partial summary as input at each step. There are multiple decoding strategies that can be implemented, as is most common we opted for a version of Beam Search \cite{vijayakumar2016diverse} with 4 beams as well as a length and repetition penalty.
Models are optimised with mini-batched AdamW \cite{loshchilov2018decoupled} with weight decay and a learning rate of $2e^{-5}$ including a linear warm-up over the first $2\%$ of steps followed by a cosine decay schedule \cite{he2019bag}. The total number of steps is fixed at 100,000 resulting in a potentially different number of epochs per model depending on the number of articles. For computing the ROUGE-1 score, we used the python version of \texttt{rouge-score 0.1.2.}\footnote{https://pypi.org/project/rouge-score/}.

\subsubsection{Content Violation Detection and Explanation}

\begin{figure}
\centering
\includegraphics[width=0.5\textwidth]{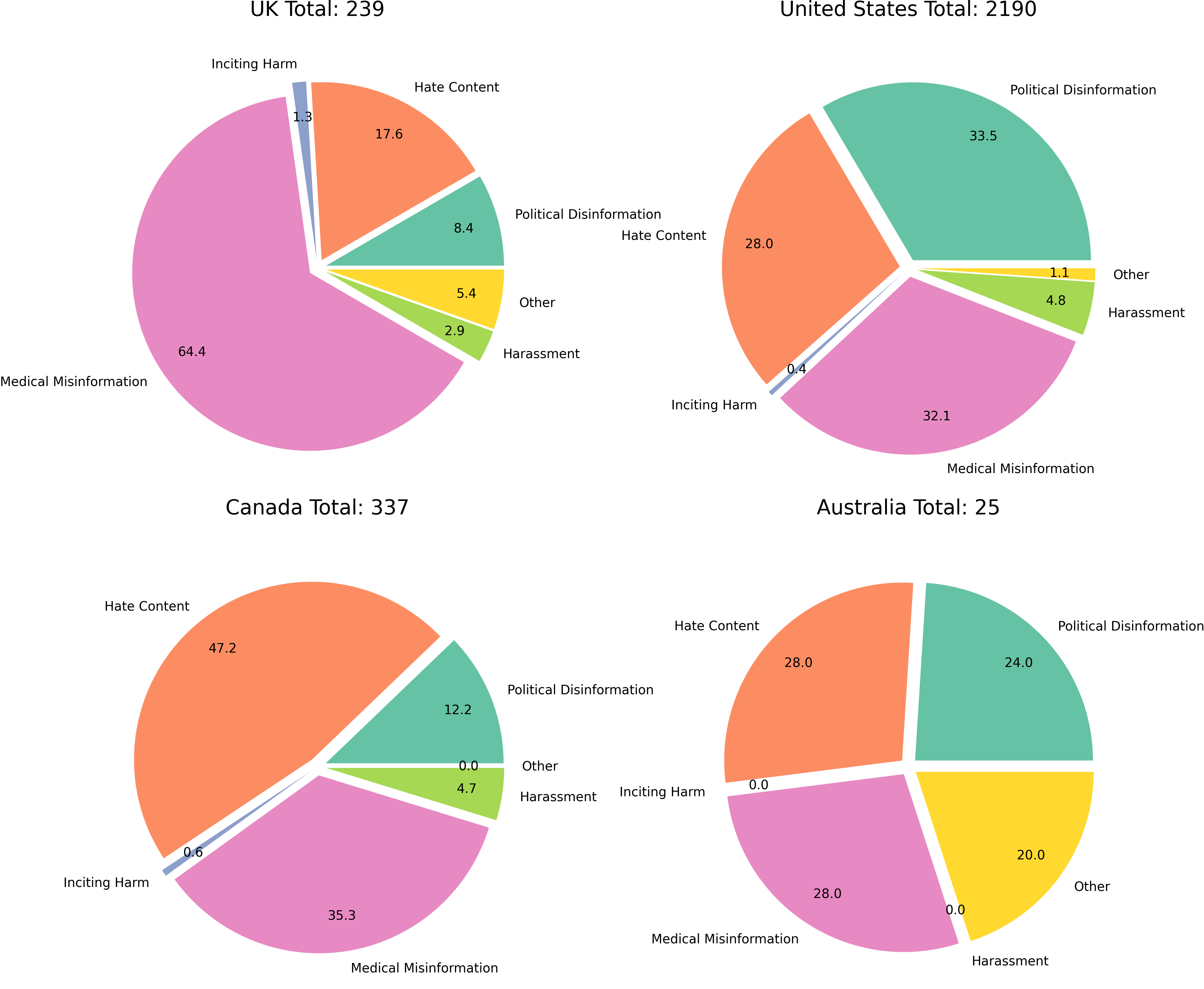}
\caption{\label{fig:violative_categories} Violative content positive examples categories breakdown.}
\end{figure}

\begin{figure*}
\centering
\includegraphics[width=0.95\textwidth]{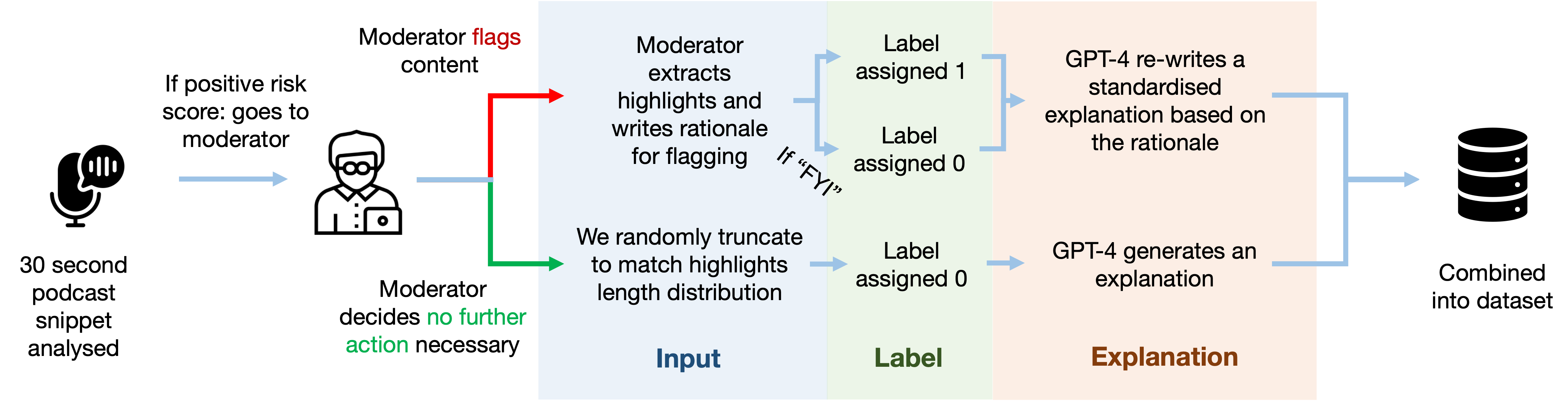}
\caption{\label{fig:data_processing} Outline of the data-processing workflow for the content violation and explanations.}
\end{figure*}

\paragraph{Dataset}
To evaluate the models' ability to flag inappropriate content, we constructed a labelled supervised learning dataset consisting of podcast snippets as the \emph{input}, a binary \emph{label} indicating whether a human moderator flagged the content due to potential violations, and an \emph{explanation} clarifying the reasons behind these decisions. The content samples were sourced from an internal safety platform and underwent a systematic transformation into a supervised dataset, as outlined in Figure \ref{fig:data_processing}. These content examples originate from user-uploaded podcasts and undergo automated risk scoring, where 30-second snippets are evaluated and assigned a score. When a snippet receives a positive score, indicating potential violative content, it is forwarded to a human moderator for a more detailed assessment. The moderator decides either to flag the content or determine that no further action is required. If the content is flagged, the moderator extracts highlights from the snippet, along with any other offending material within the entire podcast. Since the moderators write this free-form, we lightly pre-process this by using regular expressions to extract any content between quotation marks.
Furthermore, a brief rationale is provided to justify this decision. The moderator also categorizes the flagged content as either a true violation (labeled as 1) or an ``FYI'' lead. These FYI leads pertain to content that does \emph{not} currently violate guidelines but may indicate a new or emerging threat that the platform should be aware of during guideline reviews. Regardless of the categorization, an explanation is generated using GPT-4 to standardize the rationale's phrasing. This standardization is crucial, given that data originates from various moderators who may describe their rationale differently, varying from concise keywords to comprehensive background explanations. In cases where the moderator deems no further action necessary, the snippet is preserved. To ensure alignment in the input format, we randomly truncate the snippet's length to match the overall length distribution of positive examples. These snippets receive a label of 0. Since no rationale is provided by the moderator in such instances, we employ GPT-4 to generate a succinct explanation clarifying why the content may have been flagged but ultimately deemed non-violative.
The resulting dataset comprises $2,822$ positive violation examples and $4,393$ negative examples (which include the FYI leads mentioned). A majority of these examples, specifically $2,791$ positive and all negative instances, originate from four markets: United States, United Kingdom, Australia, and Canada.
Figure \ref{fig:violative_categories} gives an overview of the total count of positive instances used as training examples for each culture. Furthermore, it presents a breakdown of the percentage distribution of these examples across various topics.

\paragraph{Model architecture and training}
In this step, we further fine-tune the entire encoder-decoder of the culturally trained models of the previous step on the task of generating explanations of violations (See Figure \ref{fig:methodology}, second stage). This is a supervised sequence-to-sequence task and we deploy the \{\textit{content snippet}, \textit{rationale}\} pairs described above. In our experimental setup, this process comprises two distinct procedures: initially, a stratified fine-tuning process focuses on providing each culturally attuned model from stage 1 solely with violative training examples (i.e., rationales) relevant to its respective culture. Subsequently, another process involves fine-tuning each culturally attuned model from stage 1 with all available rationales.

The primary preference remains the first process, emphasizing the provisioning of culturally relevant examples to maintain the integrity of cultural pre-training. However, for the explanation results (as shown in Figures \ref{fig:cultural_explanations} and \ref{fig:annotators} (top), models trained on \emph{all} examples are used due to limited data availability per culture. This ensures that models can learn to produce coherent text despite data constraints. To mitigate impact on culture, when employing GPT-4 to standardize the explanations, the model is informed about the target culture of interest, allowing for partial adjustments to account for any discrepancies. All models are trained within the same computational budget. Optimization is performed using mini-batched AdamW with weight decay and a learning rate of 2e-5, incorporating a linear warm-up over the initial 2\% of steps followed by a cosine decay schedule.

Following the fine-tuning on generating rationales using the stratified process, we employ the embeddings generated by the fine-tuned encoder as the input to a classification head. The classification head per culture is further fine-tuned to identify whether the input is violative or not using data from each culture. 
More specifically, to produce the embeddings, we take the encoder of the media-diet model finetuned on generating rationales and pass the input through it to create a \texttt{[input\_length x 1024]} dimensional representation. We take the first 1024d vector as that corresponds to the \texttt{[CLS]} token introduced by the BERT tokenizer and is designed to learn an input level representation for classification during the model pre-training \cite{clark2019does,kenton2019bert}.
Once the embeddings are obtained, we pass them to a logistic regression model in order to predict the probability of violation. In each training iteration, a random 20\% subset of the dataset is designated as the test data, and this process is repeated for a total of 10 training iterations until convergence per model.  We report the average AUROC, accompanied by a 95\% confidence interval. Statistical tests reported are based on a two-sample Welch’s t-test \cite{welch:1947}.

\subsubsection{Standardising Explanations using GPT-4}
\paragraph{Prompt Details} The OpenAI API supports a turn-based dialogue format, allowing you to define the AI assistant's role using a \emph{System Prompt}t before initiating the conversation with a Conversation Prompt. In our approach, we consistently use the same System Prompt across all scenarios, while varying the Conversation Prompt depending on whether the example in question is violative or non-violative. The system prompt is as follows:

\begin{tcolorbox}[colframe=white]
    You are a helpful assistant focusing on aiding content moderators for podcast come up with helpful summaries of why a snippet of from the podcast may or may not have violated guidelines. Warning, some content may not be appropriate, but that is why you are needed to prevent it being spread online. The content moderator is from \texttt{\textbf{\{content\_country\}}}, but you are being used to moderate in \texttt{\textbf{\{target\_country\}}}, so if there is any reason content may be seen differently between the two regions you should focus on violations in the country you are moderating, without explaining it.
\end{tcolorbox}
An example of a conversational prompt for a violative case follows:
\begin{tcolorbox}[colframe=white]
    From a podcast, a moderator has highlighted these quotes:\texttt{\textbf{\{highlight\}}}
    
    They have given this rationale:\texttt{\textbf{\{rationale\}}}
    
    They say it is an example of \texttt{\textbf{\{boundary\}}}
    
    Please provide a couple sentence summary that will be useful.
\end{tcolorbox}

An example output of which is the following:
\begin{tcolorbox}[colframe=white,colback=blue!10]
    ``This snippet violates the guidelines due to the promotion of Medical Misinformation. The guest, an anti-vaccine doctor, perpetuates unverified and misleading theories about the mRNA vaccines, suggesting they are made from snake venom, involve gene editing and cause magnetic side effects. This not only perpetuates harmful anti-vaccine notions, but also echoes certain QAnon conspiracy theories.''
\end{tcolorbox}
This is an example of conversational prompt for a non-violative case:
\begin{tcolorbox}[colframe=white]
    From a podcast, a moderator has been shown this snippet that an automated tool identified as having a positive risk score of \texttt{\textbf{\{score\}}}: \texttt{\textbf{\{content\}}}. 
    
    They have decided to not take any action over it. Please provide a couple sentence summary that will be useful as to why it may have appeared potentially violative but ultimately was decided not to be - unless you strongly disagree with the moderators decision, in which case you should explain why. Don't reference the automated tool or moderator.
\end{tcolorbox}

An example output of which is:
\begin{tcolorbox}[colframe=white,colback=blue!10]
    ``The podcast snippet may have initially raised concern due to the use of strong language, such as referring to the vast majority of presidential candidates as "lunatics." This could be perceived as potentially offensive or disrespectful. However, the snippet was likely deemed non-violative as it appears to be a part of a political commentary or analysis, which generally includes such expressions of personal opinion. The speaker also discusses specific potential candidates, appearing to give a fair and balanced view rather than solely targeting or defaming one individual or group.''
\end{tcolorbox}

\paragraph{On the use of GPT-4}
We recognise that the use of GPT-4 raises concerns due to its potential introduction of biases whose complete nature remains incompletely understood. To address this, steps were taken in prompt design, such as providing the model with information about the target culture. Previous studies suggest that this approach improves the alignment of model outputs with geographic survey responses \cite{durmus2023towards}. However, we recommend moderators consider standardizing their explanations during the writing process as a proactive measure.

\subsection{Human Evaluation Study} 
\label{sec:annotators_study}
We curated a dataset consisting of twenty instances of violative content for each of the four main distinct cultures under investigation. For each content example, we generate four explanations: one from a model aligned with the same culture as the content, and three others from models randomly selected across different cultures, i.e. Hong Kong, Nigeria and Malaysia. To ensure a comprehensive assessment, we engage three distinct content moderators for each of the UK and US and two for Australia and Canada, that are experts for each of the cultures of interest. These moderators are presented with the twenty curated content examples and the complete set of generated explanations. The annotators are unaware of the cultural model which generated the explanations. Their task involves ordering the explanations based on how well they align with the local culture, with the most culturally representative explanation placed at the top. An example question presented to the annotators is show in Figure \ref{fig:annotators_example}. The moderators are then asked to order explanations A to D based on how well they perceive each explanation to align with or capture the essence of the local culture.

\begin{figure*}
\begin{tcolorbox}[colframe=white]
    ======== Question N =========
    
Content:

So for example with monkeypox, exclusively it's being pushed by the sodomites, and yet they're making it very clear you cannot stigmatize them. If someone were to say, I don't want to have a gay person because, you know, they are a threat to me. That is much more rooted in science with monkeypox then someone not getting the Covid shot which is negatively effective, right? But nonetheless they'd have your head in three seconds - lawsuits, it wouldn't get off the ground. You'd probably lose your board certification. And then speaking of monkeypox, I mean at any minute now, I don't even know when they're going to vote on this - I pray to God that that somehow they don't do before the August break so we can have time to galvanize opposition - but they might pass the banning traditional marriage in Red states bill.

Explanations:

A) This podcast snippet violates the guidelines due to its propagation of hate speech and discrimination. The host's remarks about the LGBTQ + community being a threat to monkeypox, the banning of traditional marriage in certain areas, and the dissemination of misinformation regarding political issues. The content is disrespectful and could incite harm or violence, hence its removal.

B) This podcast snippet violates the guidelines due to the spread of medical misinformation. The host makes unsubstantiated claims about monkeypox and the associated spread of monkeypox. They also misrepresent the discussion about homosexuality and the traditional marriage laws in Red states, which could potentially incite harm or discrimination.

C) The podcast snippet violates guidelines due to its spread of harmful stereotypes and promotion of harmful conspiracy theories. The host incites harmful behavior by suggesting that the spread of monkeypox is part of a sinister plot to control the population, which is discriminatory and inappropriate. Additionally, the host discriminates against same - sex marriage and gay marriage, contributing to a hostile environment. This content may incite hate or discrimination towards the LGBTQ community and spread misinformation.

D) The snippets from the podcast violate guidelines due to the propagation of harmful and unsupported conspiracy theories. The first violation is in the form of the banning of gaypox in the U. S., along with the promotion of a controversial bill in red and red in color. The second violation violates the guidelines by casting doubt on the safety of the Covid - 19 vaccines and undermining public health efforts. These violations are in violation of the guidelines and could potentially incite harm or unrest.
\end{tcolorbox}
\caption{An example question posed to the annotators.}
\label{fig:annotators_example}
\end{figure*}

\section{Results}
\subsection{Locally Improved Violation Detection - non-US models.}
\begin{figure}[h]
    \centering
        \begin{subfigure}{0.45\textwidth}
          \centering
          \includegraphics[width=\linewidth]{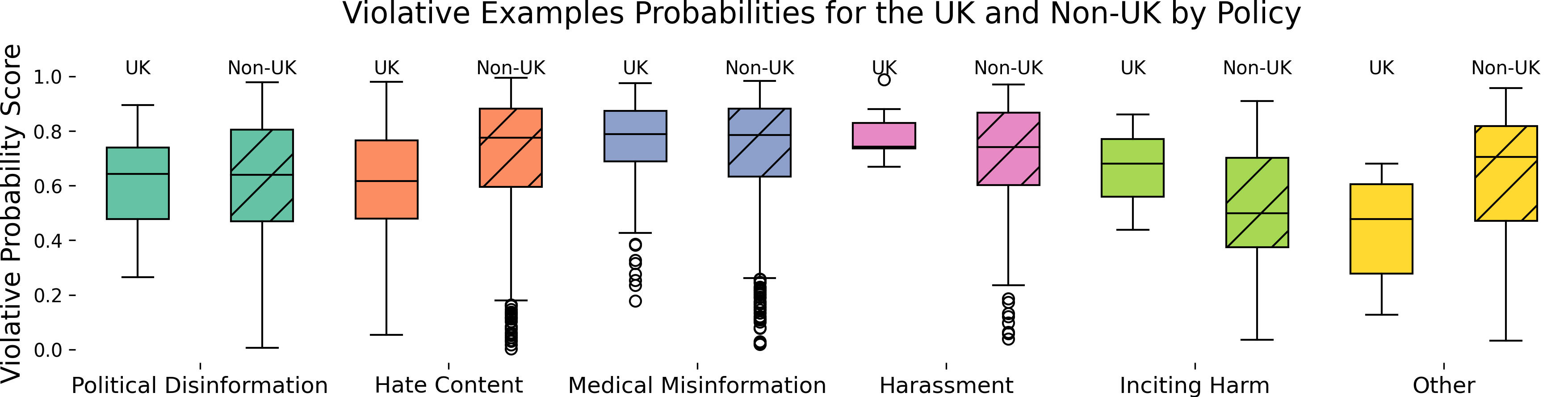}
          \caption*{UK model: score distribution on violative examples}
          \label{fig:boxplot_uk}
        \end{subfigure}
        \hspace{.2in}
        \begin{subfigure}{.45\textwidth}
          \centering
          \includegraphics[width=\linewidth]{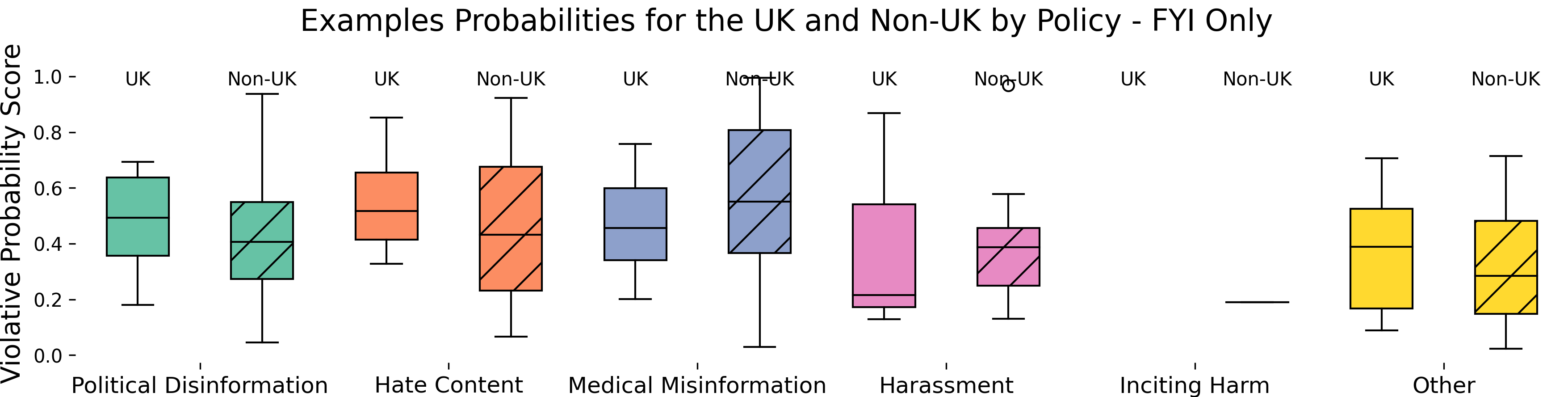}
          \caption*{UK model: FYI ony}
          \label{fig:boxplot_uk_fyi}
        \end{subfigure}
    \\
    \vspace{.5in}
    \begin{subfigure}{.45\textwidth}
      \centering
      \includegraphics[width=\linewidth]{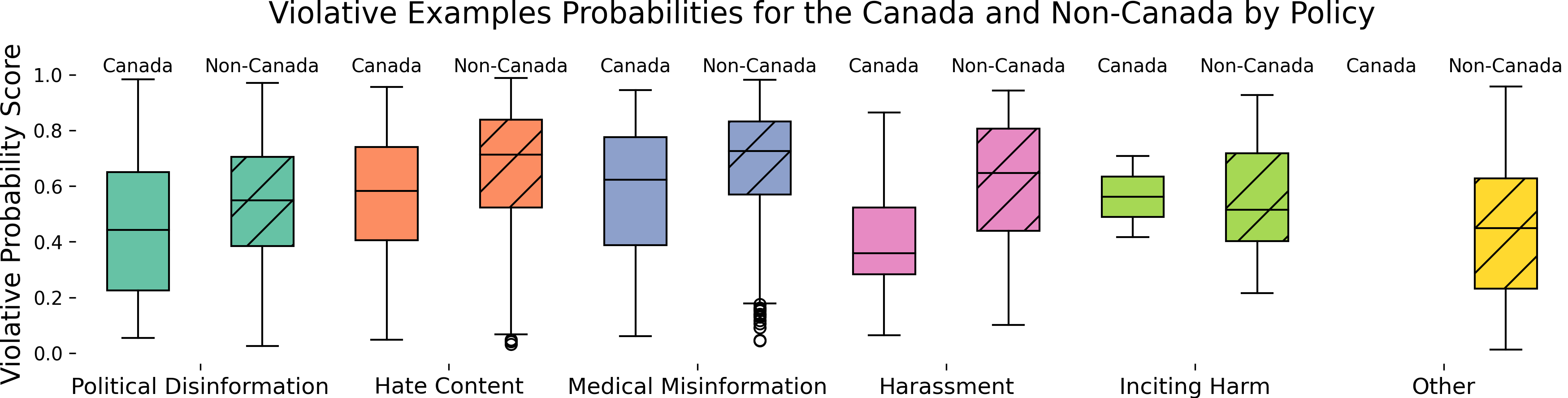}
      \caption*{CAN model: score distribution on violative examples}
      \label{fig:boxplot_can}
    \end{subfigure}
    \hspace{.2in}
    \begin{subfigure}{.45\textwidth}
      \centering
      \includegraphics[width=\linewidth]{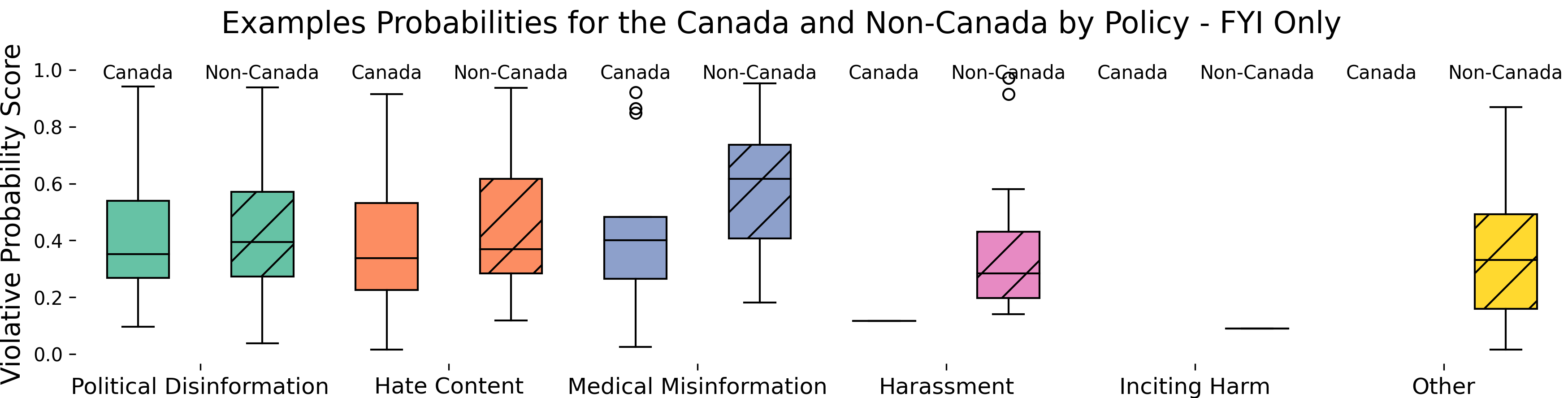}
      \caption*{CAN model: FYI only}
      \label{fig:boxplot_can_fyi}
    \end{subfigure}
    \\
    \vspace{.5in}
    \begin{subfigure}{.45\textwidth}
      \centering
      \includegraphics[width=\linewidth]{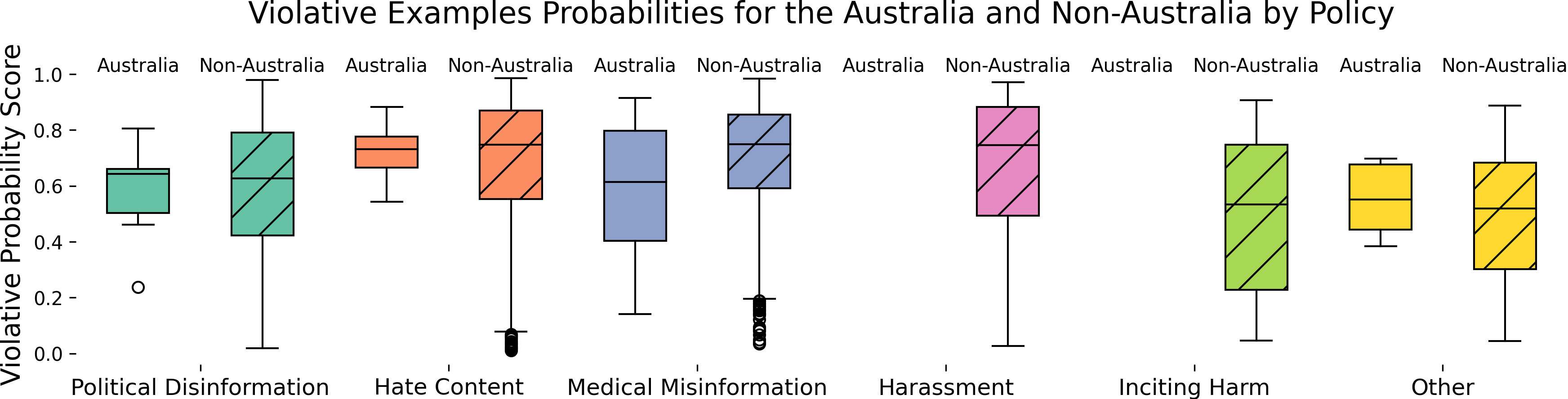}
      \caption*{AUS model: score distribution on violative examples}
      \label{fig:boxplot_aus}
    \end{subfigure}
    \hspace{.2in}
    \begin{subfigure}{.45\textwidth}
      \centering
      \includegraphics[width=\linewidth]{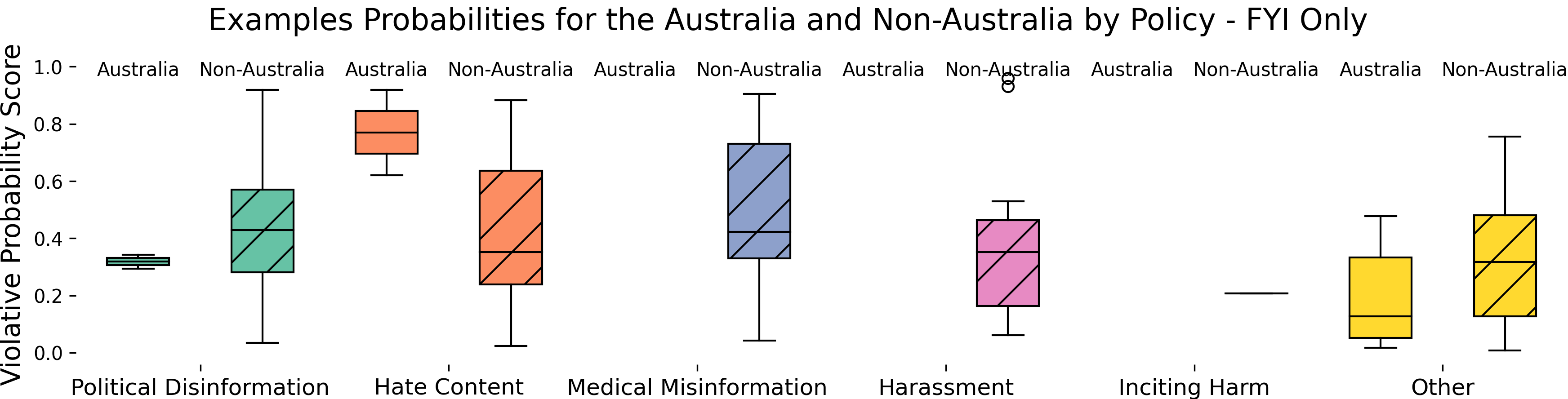}
      \caption*{AUS model: FYI only}
      \label{fig:boxplot_aus_fyi}
    \end{subfigure}
\caption{\textbf{Non-US Model probability score plots.}}
\label{fig:non_us_policy}
\end{figure}
Given that the US accumulates the largest pool of violative examples, see Figure \ref{fig:violative_categories}, it is crucial to consider this aspect in subsequent discussions. The disparity in data sizes across different cultures, notably in violative examples, complicates the task to make an analysis comparable to the one provided for the US in the main paper. The analysis of Figure \ref{fig:non_us_policy} prompts a shift in focus, pointing towards prioritizing the length of confidence intervals over absolute violation scores, particularly when evaluating cultures beyond the US. Shorter intervals within the same cultural context signify heightened prediction certainty, surpassing the significance of absolute scores.

\section{Prototype: A Tool for Assisted Moderation}
\label{sec:appendix_tool}
\begin{figure*}
\centering
\includegraphics[width=0.8\textwidth]{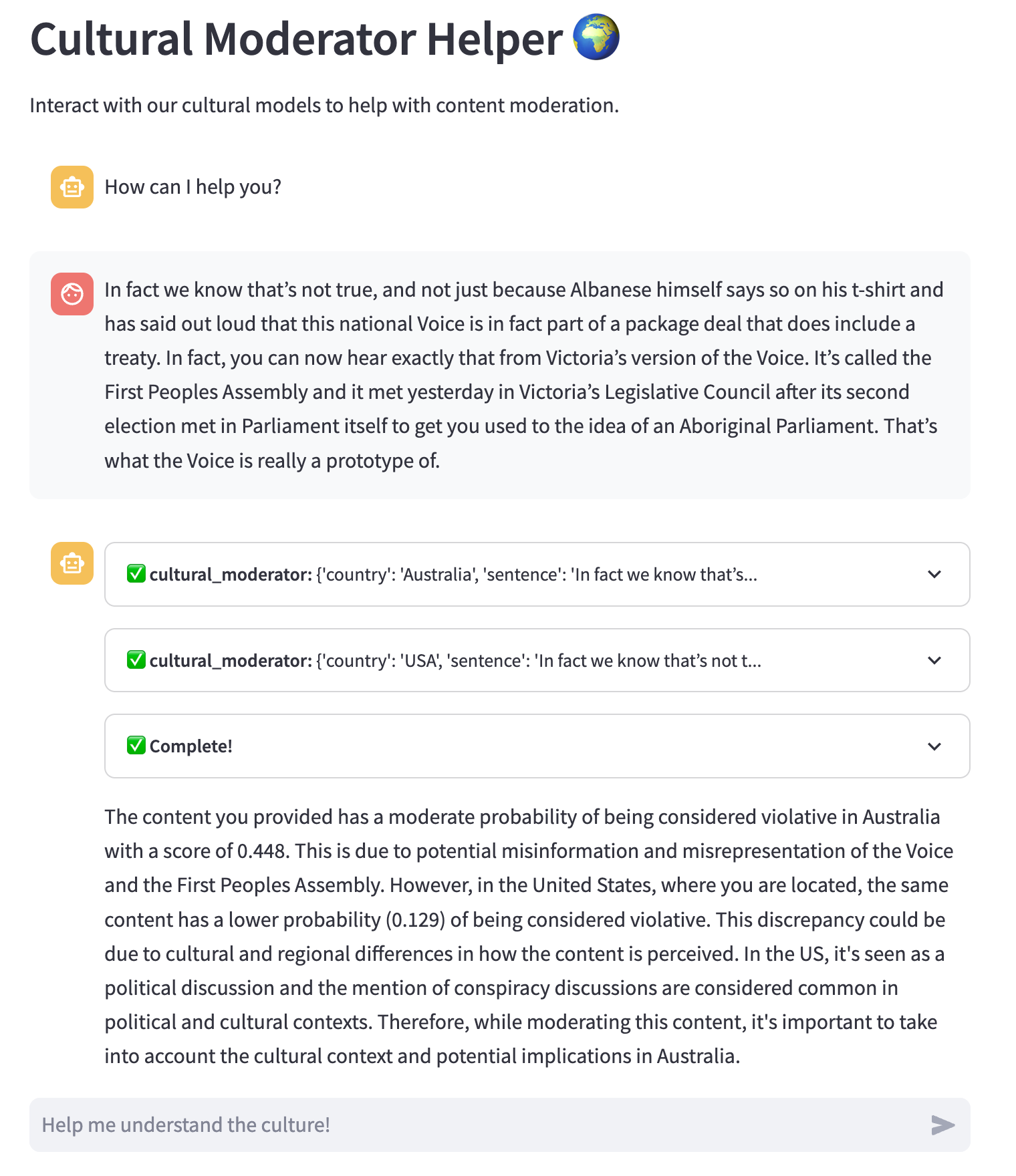}
\caption{\label{fig:tool} Assistive Moderation Tool Example - Demonstrating the annotator inputting text and requesting the model's perspective on how it might be perceived across different cultural contexts, here Australia and the USA.}
\end{figure*}

In this section, we introduce an illustrative tool that embodies our envisioned framework as an assistive technology for annotators. This tool is envisaged as an interactive platform, allowing annotators to input text they require assistance with to comprehend how it might be interpreted across diverse cultures. In the following example, the annotator inputs the text or content and requests the model's perspective on how it might be perceived in distinct cultural contexts, such as Australia and the USA, see Figure \ref{fig:tool}.

\begin{figure*}
\centering
\begin{subfigure}{\textwidth}
  \centering
  \includegraphics[width=.6\linewidth]{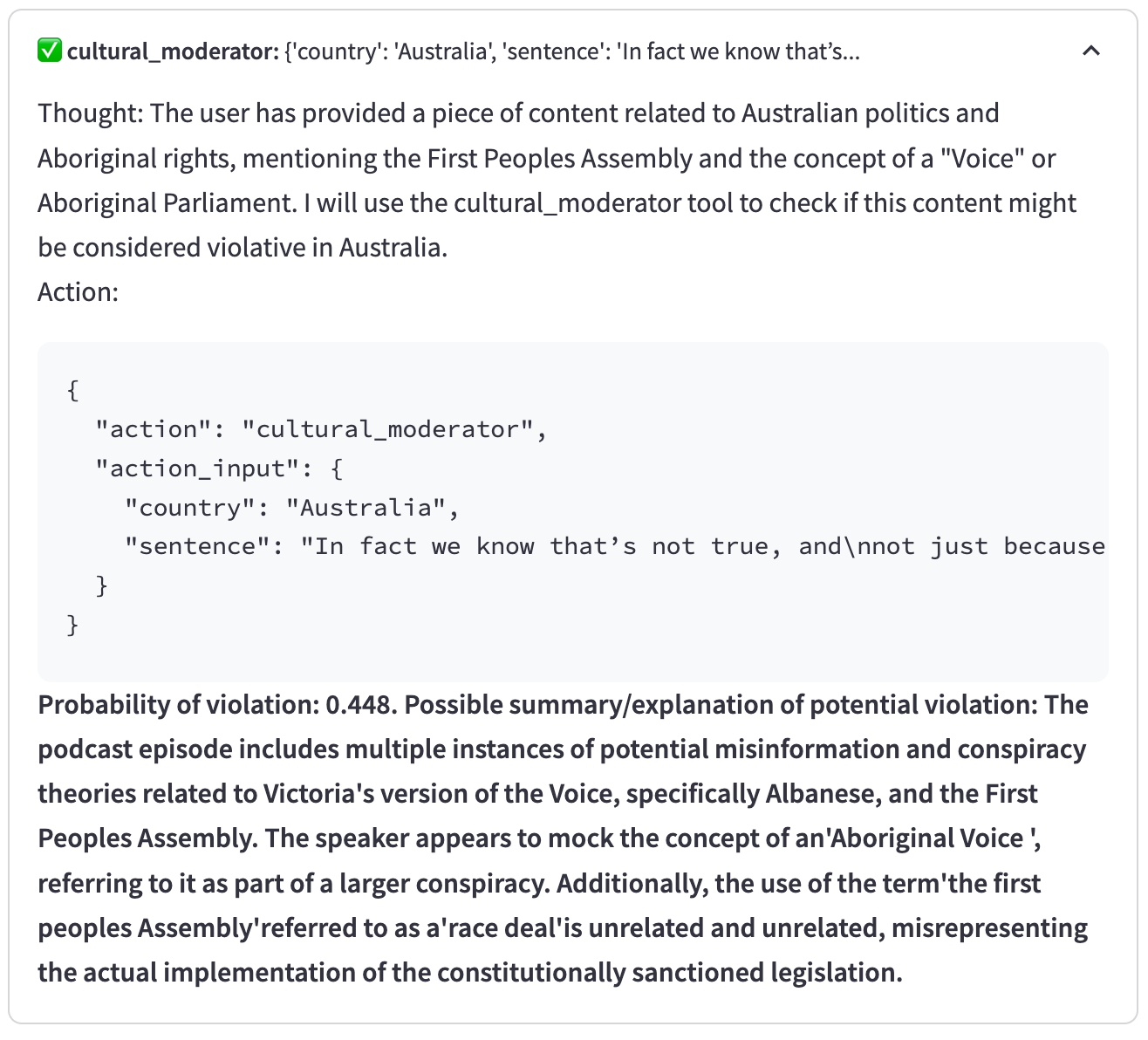}
  \caption{The Australian model output.}
  \label{fig:tool_aus}
\end{subfigure}%
\\
\begin{subfigure}{\textwidth}
  \centering
  \includegraphics[width=.6\linewidth]{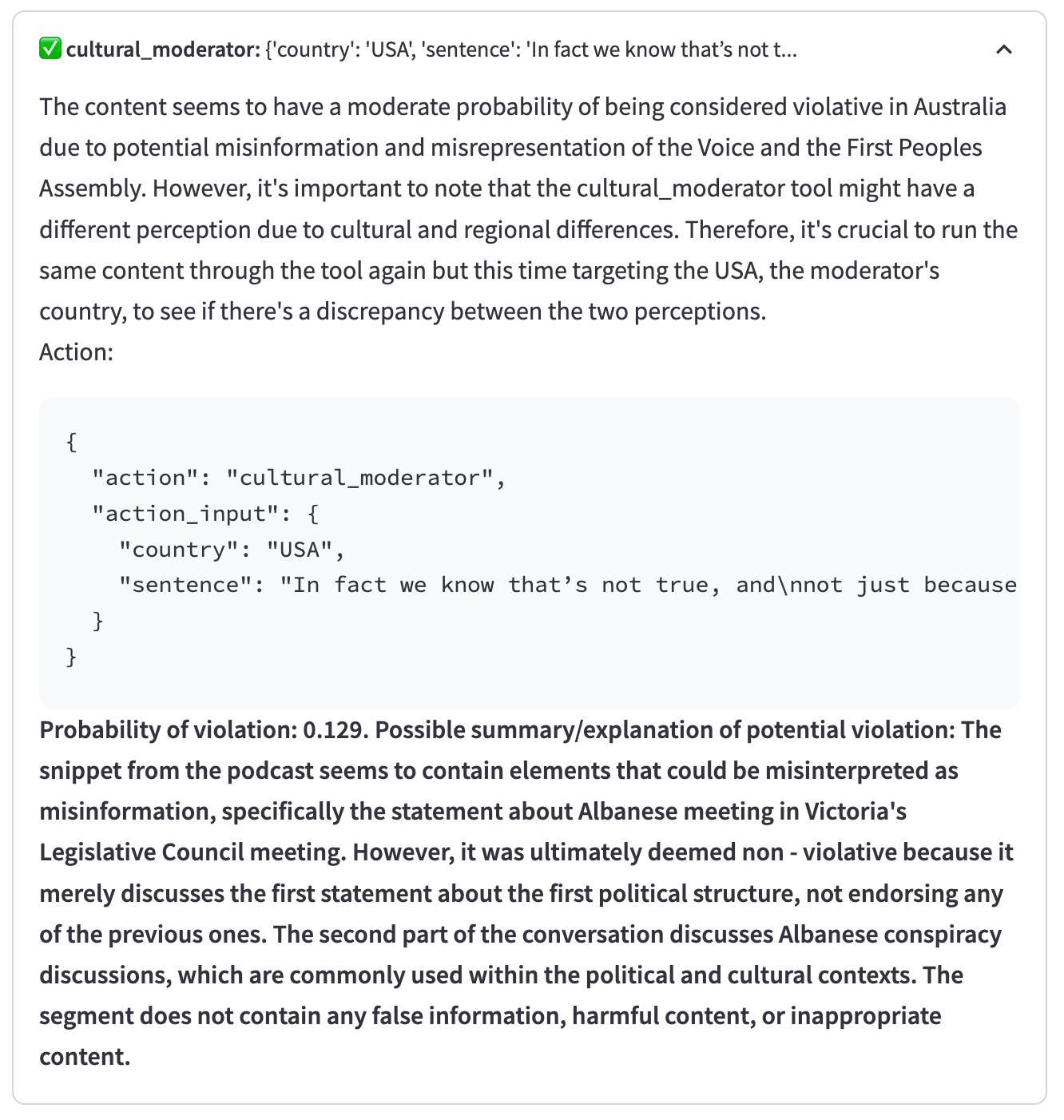}
  \caption{The US model output.}
  \label{fig:tool_usa}
\end{subfigure}
\caption{Expanded Explanations from the tool to help assist moderators.}
\label{fig:tool_expand}
\end{figure*}

The annotator can then request the tool to provide detailed rationales on how various cultures perceive the violation, as seen in Figures \ref{fig:tool_aus} and \ref{fig:tool_usa}. This facilitates a detailed understanding for the annotator regarding how different cultures may interpret the violation. It is essential to note that the annotator drives the process, utilizing the tool to gain informed perspectives.

\section{Ethical Considerations}
\label{sec:ethics}
In this work, we deploy pre-trained large language models in identifying and explain content violations in a culturally sensitive way. As such, the implementation inherits risks associated with existing biases in these models. We recognise that the automatic decision and generation of explanation of potentially violative content may be addressed as authoritative. Here, we emphasise that the present work advocates for the use of foundational models as support tools for human moderators, underscoring the necessity of human oversight. We promote this technology as a resource to provide cultural perspective for annotators to utilise as required. 

\paragraph{Dataset \& Annotators Study}
All media-diet data is sourced from private article databases (see above). Copyright and terms of service are provided by the newsapi source\footnote{https://newsapi.org/terms}. The terms of service specify that the data do not contain any personally identifiable information. Content moderation data was collected from the authors’ institution, relying on annotations provided by in-house expert annotators. All data were used for the purpose of the current research and adhere to the terms defined in their licenses.

\section{Data and Code availability}
The datasets introduced in this paper were obtained from a private vendor (news.api) and are subject to a paid license agreement. The source code and the trained models for a working version will be available at a public repository  will be made publicly available upon publication of the paper.

\section{Reproducibility Checklist}

This paper

\begin{itemize}
    \item Includes a conceptual outline and/or pseudocode description of AI methods introduced: \textbf{yes}
    \item Clearly delineates statements that are opinions, hypothesis, and speculation from objective facts and results: \textbf{yes}
    \item Provides well-marked pedagogical references for less-familiar readers to gain background necessary to replicate the paper: \textbf{yes}
\end{itemize}

\subsection{Theoretical Contributions}
Does this paper rely on one or more datasets?: \textbf{no}

\subsection{Datasets}
Does this paper rely on one or more datasets? \textbf{yes}

\begin{itemize}
    \item A motivation is given for why the experiments are conducted on the selected datasets: \textbf{yes}
    \item All novel datasets introduced in this paper are included in a data appendix:\textbf{no}. The datasets used and analyzed during the current study are available from authors upon reasonable request. 
    \item All novel datasets introduced in this paper will be made publicly available upon publication of the paper with a license that allows free usage for research purposes: \textbf{no} The datasets introduced in this paper were obtained from a private vendor (news.api) and are subject to a paid license agreement. As such, they cannot be redistributed, but researchers can access the data directly through the vendor.
    \item All datasets drawn from the existing literature (potentially including authors’ own previously published work) are accompanied by appropriate citations: \textbf{NA}
    \item All datasets drawn from the existing literature (potentially including authors’ own previously published work) are publicly available: \textbf{NA}
    \item All datasets that are not publicly available are described in detail, with explanation why publicly available alternatives are not scientifically satisfying:\textbf{NA}
\end{itemize}

\subsection{Computational Experiments}
Does this paper include computational experiments? \textbf{yes}

\begin{itemize}
    \item Any code required for pre-processing data is included in the appendix: \textbf{no}. The source code and the trained models for a working version will be available at a public repository before the camera-ready submission.
    \item All source code required for conducting and analyzing the experiments is included in a code appendix: \textbf{no}. The source code and the trained models for a working version will be available at a public repository before the camera-ready submission.
    \item All source code required for conducting and analyzing the experiments will be made publicly available upon publication of the paper with a license that allows free usage for research purposes: \textbf{yes}
    \item All source code implementing new methods have comments detailing the implementation, with references to the paper where each step comes from: \textbf{NA}
    \item If an algorithm depends on randomness, then the method used for setting seeds is described in a way sufficient to allow replication of results: \textbf{yes}
    \item This paper specifies the computing infrastructure used for running experiments (hardware and software, including GPU/CPU models; amount of memory; operating system; names and versions of relevant software libraries and frameworks: \textbf{yes}
    \item This paper formally describes evaluation metrics used and explains the motivation for choosing these metrics: \textbf{yes}
    \item This paper states the number of algorithm runs used to compute each reported result: \textbf{yes}
    \item Analysis of experiments goes beyond single-dimensional summaries of performance (e.g., average; median) to include measures of variation, confidence, or other distributional information: \textbf{yes}
    \item The significance of any improvement or decrease in performance is judged using appropriate statistical tests (e.g., Wilcoxon signed-rank): \textbf{yes}
    \item This paper lists all final (hyper-)parameters used for each model/algorithm in the paper’s experiments: \textbf{yes}
    \item This paper states the number and range of values tried per (hyper-)parameter during development of the paper, along with the criterion used for selecting the final parameter setting: \textbf{partial}
\end{itemize}

\end{document}